%% file: iclr2027_conference.tex
\documentclass{article} %
\usepackage{iclr2027_conference,times}

\input{math_commands.tex}

\usepackage{url}
\usepackage{booktabs}
\usepackage{graphicx}
\usepackage{capt-of}
\usepackage{float}
\usepackage{placeins}
\usepackage{hyperref}
\usepackage{enumitem}
\usepackage{wrapfig}

\usepackage{xspace}
\usepackage{tcolorbox}
\definecolor{oc-gray-0}{HTML}{F8F9FA}
\definecolor{oc-gray-6}{HTML}{868E96}
\newtcolorbox[auto counter]{finding}{colback=oc-gray-0,colframe=oc-gray-6,
  boxrule=0.5pt,arc=2pt,left=5pt,right=5pt,top=3pt,bottom=3pt,
  before upper={\textbf{Finding~\thetcbcounter.}\ }}

\definecolor{method-color}{HTML}{42689B}

\newcommand{\ours}{\textcolor{method-color}{\textsc{ExploreNet}}\xspace}

\title{\textsc{ExploreNet}: Learning Where to Explore\\in Diffusion GRPO}

\iclrfinalcopy
\author{Shuyue Stella Li,\quad Xiaochuang Han,\quad  Yulia Tsvetkov,\quad  Luke Zettlemoyer \\  \vspace{1mm}
University of Washington\\ \vspace{1mm}
\texttt{stelli@cs.washington.edu} \\ \vspace{1mm}
\parbox{0.03\textwidth}{\includegraphics[width=\linewidth]{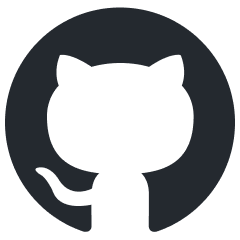}}\hspace{0.5mm}\href{https://github.com/stellalisy/ExploreNet}{\texttt{https://github.com/stellalisy/ExploreNet}}\\
}

\begin{document}
\maketitle

\input{sec/0_abstract}
\input{sec/1_intro}
\input{sec/2_background}
\input{sec/2_learned_noise_policy}
\input{sec/3_experiments}
\input{sec/4_results}
\input{sec/1_related_work}
\input{sec/5_conclusion}

\bibliography{iclr2027_conference}
\bibliographystyle{iclr2027_conference}

\appendix
\input{sec/X_suppl}

\end{document}

%% file: math_commands.tex
\usepackage{amsmath,amsfonts,bm}

\def\eqref#1{equation~\ref{#1}}

\def\1{\bm{1}}

\DeclareMathAlphabet{\mathsfit}{\encodingdefault}{\sfdefault}{m}{sl}
\SetMathAlphabet{\mathsfit}{bold}{\encodingdefault}{\sfdefault}{bx}{n}

%% file: sec/0_abstract.tex
\begin{abstract}
Group-relative RL methods such as Flow-GRPO post-train image generators by exploring with isotropic Gaussian noise added at every denoising step.
This noise decides which rollouts the model learns from, yet it perturbs every channel and spatial position of the latent equally.
In this paper, we instead show that latent elements differ in how much they change the generated image, so exploration should adapt to these differences.
We introduce \ours to learn an adaptive exploration distribution.
\ours is a policy that predicts a noise scale for every latent element from the current latent, the denoising step, and the prompt, before any reward is observed; it is trained on the reward spread of each rollout group and discarded after training, leaving inference unchanged.
On Stable Diffusion 3.5 Medium, \ours improves held-out GenEval2 by $14\%$ over Flow-GRPO, transfers to two independent compositional benchmarks and five preference and image-quality models, and reaches a $67.2\%$ human preference win-rate.
Overall, across our group-relative diffusion RL experiments, we find that exploration is learnable, the shape of the exploration distribution outweighs its magnitude, and rollout quality is more effective than rollout quantity.
\end{abstract}

%% file: sec/1_intro.tex
\section{Introduction}
\label{sec:intro}

Reinforcement learning unleashes the potential of image generators for alignment \citep{black2023ddpo,fan2023dpok}, from instruction following \citep{verifiablevisualrewards2026} and human preference \citep{xue2025dancegrpo}, to personalization \citep{dang2025personalized} and reasoning \citep{jiang2025t2ir1}.
GRPO is a common online RL algorithm that
samples a group of rollouts for each prompt and learns from the differences in their rewards
\citep{shao2024deepseekmath}.
Flow-GRPO enables GRPO in flow-matching models by adding isotropic Gaussian noise at every otherwise deterministic denoising step~\citep{liu2025flowgrpo,xue2025dancegrpo}.
This noise alone decides the rollouts and thereby provides the exploration that guides the reinforcement learning.

One line of work targets the noise,
showing that the isotropic noise injection of Flow-GRPO results in excessive or insufficient stochasticity, and proposing to adjust noise levels by denoising steps \citep{deng2026densegrpo, li2025mixgrpo}.
Within each step, however, the same noise scale still applies to every channel and spatial position of the latent.
Our key insight is that \emph{the latent is unevenly sensitive}.
Perturbing some latent channels restructures the scene, while perturbing others changes only details (Section~\ref{sec:structure}).
\textbf{Exploration should therefore adapt to how sensitive each latent element is}, yet isotropic noise allocates it equally over fine details and scene structure.

Another line of work directly manipulates rollout groups for a better learning signal by oversampling rollouts, observing their rewards, and training on an informative subset \citep{ge2025progrpo,yu2025smartgrpo}.
However,
post-hoc rollout selection relies on knowing the reward \emph{before} choosing which rollouts to train on,
and each update pays for rollouts it discards.
Instead, we frame \textbf{learning the exploration distribution as a reinforcement learning problem}.
The noise drawn at each denoising step is an action, and its reward arrives when the image is scored \emph{after} the last denoising step.

\begin{figure}[t]
    \centering
    \includegraphics[width=0.98\linewidth]{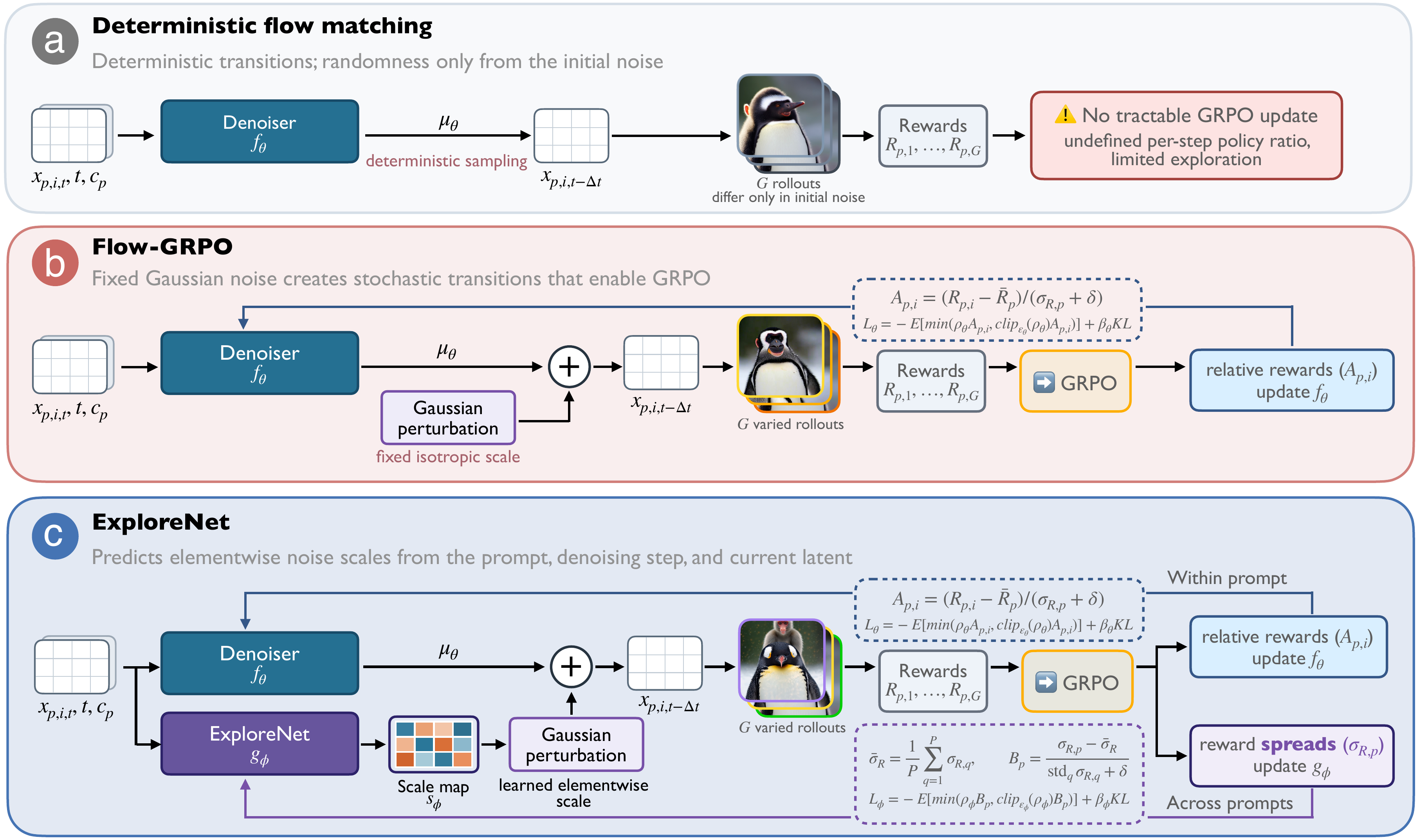}\vspace{-3mm}
    \caption{%
\hspace{-1mm}\parbox{0.025\textwidth}{\includegraphics[width=\linewidth]{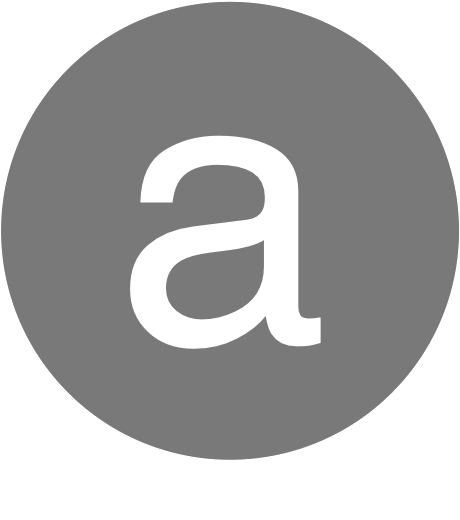}}\hspace{-0.5mm}
Deterministic flow-matching sampling adds no randomness beyond the initial noise, which limits exploration and leaves the policy ratio without a tractable transition density.
\hspace{-0.5mm}\parbox{0.025\textwidth}{\includegraphics[width=\linewidth]{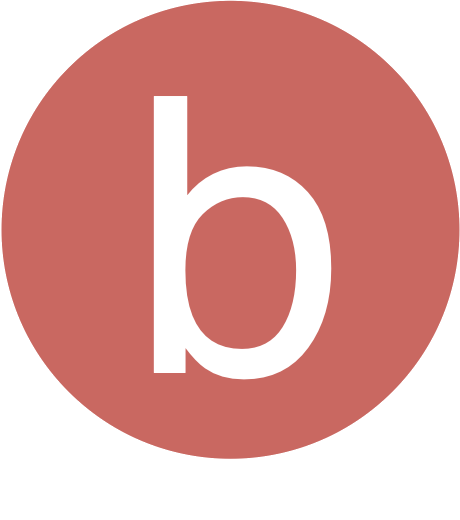}}\hspace{-0.5mm}
Flow-GRPO samples rollouts with isotropic Gaussian perturbations and updates the denoiser $f_\theta$ with group-relative advantages $A_{p,i}$.
\hspace{-0.5mm}\parbox{0.025\textwidth}{\includegraphics[width=\linewidth]{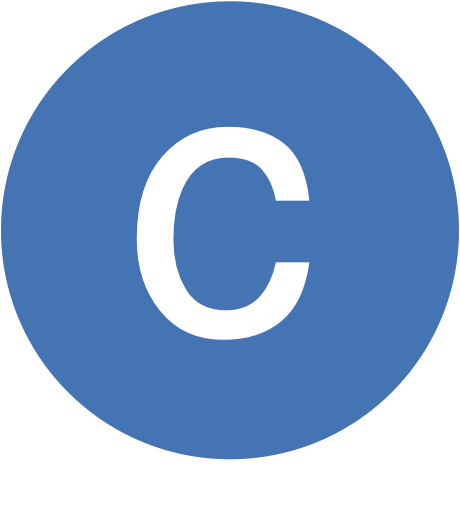}}\hspace{-0.5mm}
\ours $g_\phi$ predicts the scale map $s_\phi$ before each denoising step.
The denoiser keeps $\mathcal{L}_\theta$, and \ours optimizes $\mathcal{L}_\phi$ with the normalized reward spread $B_p$.
Inference uses only the trained denoiser.}
    \label{fig:explorenet-overview}\vspace{-3mm}
\end{figure}

More specifically, we propose \ours, a diffusion post-training method that learns an adaptive noise scale for each latent element, sets it before each denoising step without observing any reward, and keeps every rollout it samples for training.
\ours is a separate transformer network that predicts noise scales
from the current latent, the denoising step, and the prompt (Figure~\ref{fig:explorenet-overview}), and is trained jointly with the denoiser on the reward spread of each prompt group.
It is removed after training, so the trained denoiser keeps its sampler and inference cost.

On Stable Diffusion 3.5 Medium \citep[SD3.5-M]{esser2024scaling} trained with GenEval2 \citep{kamath2025geneval2}, \ours outperforms Flow-GRPO across a broad image-generation evaluation suite.
Learning the exploration distribution improves held-out GenEval2 by $14\%$, and the gain transfers to two independent compositional benchmarks and five preference and image-quality models.
Human raters prefer \ours to Flow-GRPO and the pretrained checkpoint $67.2\%$ of the time.
Further ablations show that the gain comes from the \emph{shape} of the exploration distribution.
Noise raised uniformly to \ours's average variance performs worse than Flow-GRPO, and \ours's scales are largest in the latent channels whose perturbation changes the image most.
Rollout quality is also more effective than rollout quantity.
With 8 rollouts per prompt, \ours exceeds Flow-GRPO with 24, and with half as many rollouts, it matches privileged post-hoc selection, which observes every reward.
Our contributions are as follows:

\vspace{-1mm}
\begin{itemize}[itemsep=1pt,topsep=0pt,leftmargin=15pt]
\item \ours, which learns a noise scale for every latent element at each denoising step, set from the current latent, the denoising step, and the prompt before any reward is observed, at unchanged inference cost.
\item Experiments on SD3.5-M showing that exploration is learnable, with gains over Flow-GRPO across a broad image-generation evaluation suite and in human preference.
\item Controlled comparisons showing that the shape of the exploration distribution outweighs its magnitude and that rollout quality is more effective than rollout quantity.
\end{itemize}

%% file: sec/2_background.tex
\vspace{-1mm}
\section{Preliminaries: Group-Relative RL for Flow Models}
\label{sec:background}\vspace{-1mm}

Following DDPO and Flow-GRPO, we treat denoising as a Markov decision process \citep{black2023ddpo,liu2025flowgrpo}.
At denoising step $t$, the state consists of the latent $x_t$, the step $t$, and the prompt $c$; the action is the next latent $x_{t-\Delta t}$; and the reward is zero at every step except the last, where the finished image receives $R$.
Because a flow-matching transition is deterministic, Flow-GRPO samples the action from a Gaussian centered on the deterministic update, which supplies exploration and a tractable transition density.

As shown in Figure~\ref{fig:explorenet-overview}b, each training iteration samples $P$ prompt groups.
Prompt group $p$ has text conditioning $c_p$ and $G$ rollouts indexed by $i$.
We write $x_{p,i,t}$ for the latent at denoising step $t$ and $R_{p,i}$ for its terminal reward.
Let $f_\theta$ denote the denoiser, a diffusion transformer, and $\mu_\theta(x_{p,i,t},t,c_p)$ the transition mean it induces.
With the noise coefficient $a_t=\sigma_t\sqrt{|\Delta t|}$, where $\sigma_t=\lambda\sqrt{t/(1-t)}$ and $\lambda$ is a fixed noise-level hyperparameter, Flow-GRPO samples
\begin{equation}
  x_{p,i,t-\Delta t} = \mu_\theta(x_{p,i,t},t,c_p)+a_t z_{p,i,t}, \qquad z_{p,i,t}\sim\mathcal{N}(0,I).
  \label{eq:flowgrpo-sde}
\end{equation}
The noise scale at a given denoising step is therefore the same for every prompt, latent, and latent element.
Latent elements differ, however, in how much a perturbation changes the generated image (Section~\ref{sec:structure}),
and a shared scale cannot adapt exploration to these differences.

Flow-GRPO normalizes terminal rewards within each prompt group:
\begin{align}
  \bar R_p &= \frac{1}{G}\sum_{i=1}^{G}R_{p,i},
  &
  \sigma_{R,p}
  &=
  \left(\frac{1}{G}\sum_{i=1}^{G}(R_{p,i}-\bar R_p)^2\right)^{1/2},
  &
  A_{p,i}
  &=
  \frac{R_{p,i}-\bar R_p}{\sigma_{R,p}+\delta},
  \label{eq:advantage-normalization}
\end{align}
where $\sigma_{R,p}$ is the standard deviation of the rewards in group $p$, which we call its \emph{reward spread}, and $\delta$ provides numerical stability.
The terminal advantage $A_{p,i}$ is assigned to every transition in rollout $i$.
Next, let $p_\theta(x_{t-\Delta t}\mid x_t,t,c)=\mathcal{N}(\mu_\theta,\,a_t^2 I)$ denote the Gaussian transition density of Eq.~\ref{eq:flowgrpo-sde}.
For data collected by $\theta_{\mathrm{old}}$, the Flow-GRPO denoising ratio and clipped loss are
\begin{equation}
  \rho_{\theta,p,i,t}
  =
  \frac{ p_\theta(x_{p,i,t-\Delta t}\mid x_{p,i,t},t,c_p) }{ p_{\theta_{\mathrm{old}}}(x_{p,i,t-\Delta t}\mid x_{p,i,t},t,c_p) },
  \label{eq:denoising-policy-ratio}
\end{equation}
\begin{equation}
  \mathcal{L}_\theta
  =
  -\mathbb{E}_{p,i,t} \left[ \min\left( \rho_{\theta,p,i,t}A_{p,i}, \operatorname{clip}_{\epsilon_\theta} (\rho_{\theta,p,i,t})A_{p,i} \right) \right]+\beta_\theta\,\mathbb{E}_{p,i,t}\big[D_{\mathrm{KL}}(p_\theta\,\|\,p_{\mathrm{ref}})\big],
  \label{eq:denoising-policy-loss}
\end{equation}
where $\operatorname{clip}_{\epsilon}(\rho)$ truncates the ratio to $[1-\epsilon,1+\epsilon]$ and $p_{\mathrm{ref}}$ is the transition density of the pretrained denoiser.
\ours retains this denoising objective and learns the rollout noise.

%% file: sec/2_learned_noise_policy.tex
\section{\textsc{ExploreNet}}
\label{sec:learned-noise}\vspace{-1mm}

In the decision process of Section~\ref{sec:background}, each action $x_{t-\Delta t}=\mu_\theta+a_t z$ has two parts: the mean, set by the denoiser, and the noise.
\ours is a second policy whose action is the noise: before the reward is known, it predicts a scale for each latent element from the current latent, the denoising step, and the prompt, and it is trained afterward from the reward spread of each rollout group.
Figure~\ref{fig:explorenet-overview}c shows how the two policies interact during training.

\subsection{Predicting Elementwise Noise Scales}\vspace{-1mm}
A separate scale for each latent element lets exploration selectively concentrate on different elements, which the shared scale of Eq.~\ref{eq:flowgrpo-sde} cannot do.
Because the noise must be set before each step, each scale can depend only on the inputs the denoiser receives at that step, namely the current latent, the denoising step, and the prompt.
A diffusion transformer takes exactly these inputs and returns one output per latent element, so we parameterize \ours as a second SD3.5-M transformer $g_\phi$, initialized from the pretrained weights and trained through LoRA adapters.
At each denoising step, it maps the current state to one score per latent element:
\begin{equation}
  h_{\phi,p,i,t}=g_\phi(x_{p,i,t},t,c_p),
\end{equation}
where $h_{\phi,p,i,t,j}$ is the score of latent element $j$, indexed over the channel and spatial dimensions of the latent.
We standardize these scores within each latent and convert them to bounded log-scales:\vspace{-9mm}

\begin{align}
  \widetilde h_{\phi,p,i,t,j}
  &=
  \frac{ h_{\phi,p,i,t,j}-\operatorname{mean}_{j'}h_{\phi,p,i,t,j'} }{ \max\!\left(\operatorname{std}_{j'}h_{\phi,p,i,t,j'},\delta_h\right) },
  \label{eq:explorenet-normalize}\\
  \ell_{\phi,p,i,t,j}
  &=
  \operatorname{clip} \left(\widetilde h_{\phi,p,i,t,j},\ell_{\min},\ell_{\max}\right), \qquad s_{\phi,p,i,t,j}=\exp(\ell_{\phi,p,i,t,j}).
  \label{eq:explorenet-clamp}
\end{align}
Standardization fixes the mean and standard deviation of each latent's scores before clipping, so the policy cannot raise or lower all of a latent's log-scales together, and a larger scale on some elements comes with a smaller scale on others.
Clipping then bounds every scale to $[e^{\ell_{\min}},e^{\ell_{\max}}]$.
The policy therefore sets the \emph{shape} of the exploration distribution, how noise is allocated across latent elements.
Its \emph{magnitude}, the average noise variance $\operatorname{mean}_j s_{\phi,p,i,t,j}^2$, has no separate control: it is determined by that shape and is at most $e^{2\ell_{\max}}$.

Let $s_{\phi,p,i,t}$ denote the \emph{scale map}, the vector of scales $s_{\phi,p,i,t,j}$ over all latent elements.
\ours samples
\begin{equation}
\begin{aligned}
  z_{p,i,t} &\sim \mathcal{N}(0,I),
  &\eta_{p,i,t} &= s_{\phi,p,i,t}\odot z_{p,i,t}, \\
  q_\phi(\eta_{p,i,t}\mid x_{p,i,t},t,c_p)
    &= \mathcal{N}\!\left(0,\operatorname{diag}(s_{\phi,p,i,t}^{2})\right),
  &x_{p,i,t-\Delta t}
    &= \mu_\theta(x_{p,i,t},t,c_p)+a_t\eta_{p,i,t}.
\end{aligned}
  \label{eq:explorenet-transition}
\end{equation}
The denoiser sets the transition mean and \ours reshapes the noise around it, so rollouts come from an exploration distribution centered on the denoiser's trajectories.
The denoiser keeps the objective of Eq.~\ref{eq:denoising-policy-loss}, and its ratio $\rho_\theta$ (Eq.~\ref{eq:denoising-policy-ratio}) evaluates each sampled transition under the isotropic density of Eq.~\ref{eq:flowgrpo-sde} for both $\theta$ and $\theta_{\mathrm{old}}$.
The transitions themselves come from \ours, so the score $\nabla_{\mu_\theta}\log p_\theta=(x_{p,i,t-\Delta t}-\mu_\theta)/a_t^2=s_{\phi,p,i,t}\odot z_{p,i,t}/a_t$ weights each latent element by its noise scale.
\ours therefore shapes both which transitions the denoiser learns from and how strongly each latent element contributes to the denoiser's update.

\subsection{Training from Reward Spread}

\begin{wrapfigure}{r}{0.45\textwidth}
\vspace{-5.4mm}
\centering
\includegraphics[width=\linewidth]{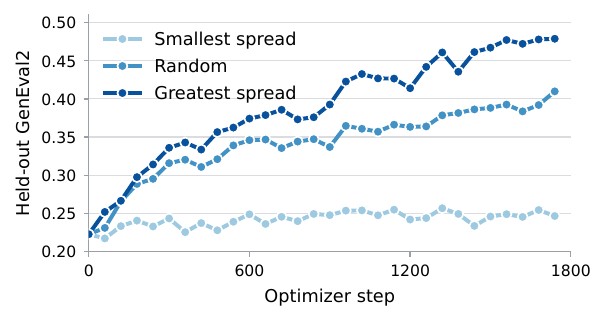}
\vspace{-9mm}
\caption{With all else fixed, selecting high-spread groups yields the fastest learning.}
\label{fig:reward-spread-selection}
\vspace{-2mm}
\end{wrapfigure}

A noise action has no reward of its own: the reward arrives only for the finished image, and the denoiser learns from the reward differences among rollouts of the same prompt.
A noise choice can therefore be credited only through the group it helps produce, and what a group offers the denoiser is its reward spread.
A group whose rewards are equal gives every rollout zero advantage, and Flow-GRPO divides each group's advantages by its reward spread $\sigma_{R,p}$ (Eq.~\ref{eq:advantage-normalization}), so a group whose rewards are nearly equal yields advantages that amplify small, uninformative differences.
To measure the effect of spread directly, we generate 16 rollouts per prompt and train on 8 selected to have the smallest reward spread, selected at random, or selected to have the greatest reward spread, with the rollout pool, update group size, and optimizer fixed.
Greatest-spread selection learns fastest and smallest-spread selection slowest (Figure~\ref{fig:reward-spread-selection}), and the ordering holds on eight of nine evaluation metrics (Appendix~\ref{sec:supp-spread-selection}).
We therefore take the reward spread $\sigma_{R,p}$ of each group (Eq.~\ref{eq:advantage-normalization}) as the return of \ours and standardize it across the $P$ prompt groups of an iteration,
\begin{equation}
  B_p = \frac{\sigma_{R,p}-\operatorname{mean}_{q}\sigma_{R,q}}{\operatorname{std}_{q}\sigma_{R,q}+\delta},
  \label{eq:noise-advantage}
\end{equation}
where $q$ ranges over the $P$ groups.
Every sampled noise vector $\eta_{p,i,t}$ in prompt group $p$ receives the same advantage $B_p$, so training raises the probability of the noise vectors sampled in groups whose spread exceeds the average of the iteration.
For a rollout sampled with $\phi_{\mathrm{old}}$, the ratio for \ours is
\begin{equation}
  \rho_{\phi,p,i,t} = \frac{ q_\phi(\eta_{p,i,t}\mid x_{p,i,t},t,c_p) }{ q_{\phi_{\mathrm{old}}}(\eta_{p,i,t}\mid x_{p,i,t},t,c_p) }.
  \label{eq:noise-policy-ratio}
\end{equation}
We optimize the clipped exploration objective
\begin{equation}
\begin{aligned}
  \mathcal{L}_\phi
  ={}& -\mathbb{E}_{p,i,t}\left[
    \min\left( \rho_{\phi,p,i,t}B_p, \operatorname{clip}_{\epsilon_\phi} (\rho_{\phi,p,i,t})B_p \right)
  \right] \\
  &+\beta_\phi\,\mathbb{E}_{p,i,t}\big[D_{\mathrm{KL}}(q_\phi\,\|\,\mathcal{N}(0,I))\big].
  \label{eq:noise-policy-loss}
\end{aligned}
\end{equation}
The KL term regularizes the predicted scales toward $\mathcal{N}(0,I)$, the unscaled transition noise of Flow-GRPO in Eq.~\ref{eq:flowgrpo-sde}.

\subsection{Joint Training and Inference}
Each training iteration samples $P$ prompts and $G$ rollouts per prompt using the current denoiser and \ours, and the same rewards train both policies at two granularities.
The rollout advantage $A_{p,i}$ measures each rollout's reward relative to its prompt group and weights the policy gradient of the denoiser, and the group advantage $B_p$ measures each group's reward spread relative to the other groups in the iteration and weights the policy gradient of \ours.
The two networks minimize $\mathcal{L}_\theta$ and $\mathcal{L}_\phi$ with separate optimizers.
At inference, we discard $g_\phi$ and sample the trained denoiser with the standard flow-matching sampler.

%% file: sec/3_experiments.tex
\section{Experiments}
\label{sec:experiments}

\subsection{Experimental Setup}
\label{sec:setup}

We test our central hypothesis: allocating rollout noise across latent elements improves group-relative diffusion RL.
Our experiments address three questions: (1)~Does \ours improve the trained denoiser?
(2)~Why does it work?
(3)~Where does it place the noise?

\textbf{Training.} All training experiments fine-tune SD3.5-M \citep{esser2024scaling} on the GenEval2 training split~\citep{kamath2025geneval2} with the Flow-GRPO denoising objective \citep{liu2025flowgrpo}.
Within each comparison, methods use the same training prompts, reward implementation, denoising schedule, rollout group size, optimizer settings for the denoiser, training duration, and checkpoint cadence.
Appendix~\ref{sec:supp-training-details} gives the complete training hyperparameters and computational cost.

\textbf{Baselines and controls.} We compare \ours with Flow-GRPO and two controls:\vspace{-1mm}

\begin{itemize}[itemsep=0pt,topsep=0pt,leftmargin=15pt]
    \item \textbf{Flow-GRPO} uses the isotropic noise of Eq.~\ref{eq:flowgrpo-sde} with the default noise level $\lambda=0.7$ and 24 rollouts per prompt group, the same group size as \ours.

    \item \textbf{Matched isotropic control} replaces \ours's learned scale map with a single constant, $1.41$, for every latent element, which matches the average variance of \ours's noise (Appendix~\ref{sec:supp-varmatch}).
    It removes only the allocation to test whether the gain comes from the shape or the magnitude of \ours's exploration.

    \item \textbf{Privileged post-hoc selection} generates 48 rollouts per prompt, observes every reward, and trains on the 24 with the greatest reward spread.
    It uses twice the rollouts and reward evaluations of \ours, so it tests whether learned allocation, set before any reward is observed, can match or exceed selection after sampling.
\end{itemize}

\textbf{Automatic evaluation.} On the held-out GenEval2 prompts, we score the trained models with the GenEval2 metric and with a VLM judge based on GPT-4o~\citep{openai2024gpt4o}.
To evaluate transfer performance, we use GenEval~\citep{ghosh2023geneval}, VVRBench-Fast~\citep{verifiablevisualrewards2026}, PickScore~\citep{kirstain2023pickscore}, HPSv2.1~\citep{wu2023human}, HPSv3~\citep{ma2025hpsv3}, Aesthetic~\citep{schuhmann2022aesthetic}, and ImageReward~\citep{xu2023imagereward}.
Appendix~\ref{sec:supp-evaluation-details} specifies the prompt sets, metric implementations, generation settings, and other evaluation details.

\textbf{Human evaluation.}
Three annotators each make 216 pairwise preference judgments (A, B, or Tie) over 72 unique prompts from pairs drawn from SD3.5-M, Flow-GRPO, and \ours.
Win-rate is calculated as each model's total score divided by the number of comparisons it appears in (win receives a score of $1$, tie receives a score of $0.5$).
The prompts are sampled equally from DrawBench~\citep{saharia2022imagen}, PartiPrompts~\citep{yu2022parti}, DPG-Bench~\citep{hu2024ella}, T2I-CompBench~\citep{huang2023t2icompbench}, the OCR prompt set from Flow-GRPO~\citep{liu2025flowgrpo}, and VVRBench~\citep{verifiablevisualrewards2026}.
Appendix~\ref{sec:supp-human-evaluation} gives the annotation protocol details.

%% file: sec/4_results.tex
\subsection{\ours Improves the Trained Denoiser}
\label{sec:results-main}

\begin{figure}[t]
\centering
\begin{minipage}[t]{0.53\linewidth}
    \vspace{-3mm}
    \captionof{table}{Final-checkpoint results, mean and sample standard deviation over three training seeds.
Appendix~\ref{sec:supp-per-seed} reports each seed.}
    \label{tab:main-results}\vspace{1.5mm}
    \centering
    \small
    \setlength{\tabcolsep}{2.0pt}
    \renewcommand{\arraystretch}{1.05}
    \makebox[\linewidth][c]{%
    \resizebox{\linewidth}{!}{
    \begin{tabular}{@{}lrrrr@{}}
    \toprule
    Metric & \hspace{-3mm}Pretrained & Flow-GRPO & \ours & $\Delta$ \\
    \midrule
    GenEval2 & $0.2375$ & $0.4957_{\pm0.0025}$ & $\mathbf{0.5661_{\pm0.0318}}$ & $+0.0704$ \\
    VLM Judge & $4.0250$ & $4.5083_{\pm0.0946}$ & $\mathbf{4.7292_{\pm0.0402}}$ & $+0.2208$ \\
    GenEval & $0.6157$ & $0.6955_{\pm0.0095}$ & $\mathbf{0.7350_{\pm0.0096}}$ & $+0.0395$ \\
    VVR-Fast & $0.756$ & $0.7646_{\pm0.0062}$ & $\mathbf{0.7902_{\pm0.0188}}$ & $+0.0256$ \\
    PickScore & $0.8405$ & $0.8448_{\pm0.0019}$ & $\mathbf{0.8512_{\pm0.0006}}$ & $+0.0064$ \\
    HPSv2.1 & $0.3001$ & $0.2993_{\pm0.0023}$ & $\mathbf{0.3056_{\pm0.0040}}$ & $+0.0063$ \\
    HPSv3 & $7.6892$ & $8.2112_{\pm0.1784}$ & $\mathbf{8.8043_{\pm0.1535}}$ & $+0.5931$ \\
    Aesthetic & $5.5165$ & $5.4960_{\pm0.0261}$ & $\mathbf{5.6638_{\pm0.0596}}$ & $+0.1678$ \\
    ImageReward & $0.9291$ & $1.1137_{\pm0.0263}$ & $\mathbf{1.2774_{\pm0.0154}}$ & $+0.1637$ \\
    \bottomrule
    \end{tabular}%
    } }
    \raggedright
\end{minipage}\hfill
\begin{minipage}[t]{0.45\linewidth}
    \vspace{3mm}
    \centering
    \includegraphics[width=\linewidth]{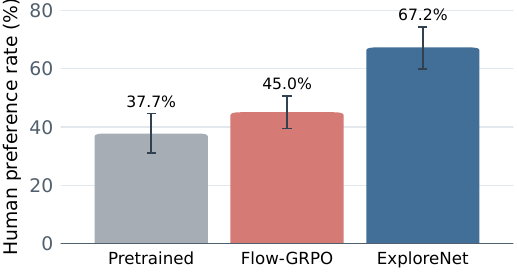}\vspace{-4mm}
    \raggedright
    \captionof{figure}{Human preference win-rate across three annotators on 216 comparisons; avg. pairwise agreement is 64.4\%.}
    \label{fig:human-final-outputs}\vspace{-3mm}
\end{minipage}
\end{figure}

We begin by comparing \ours with Flow-GRPO, with the unmodified SD3.5-M as a pretrained reference.
Across three matched training seeds, \textbf{\ours outperforms Flow-GRPO on every metric in the evaluation suite}  (Table~\ref{tab:main-results}).
The GenEval2 gain persists across the final five checkpoints and under a paired bootstrap over evaluation prompts (Appendix~\ref{sec:supp-prompt-bootstrap}).

\begin{figure*}[t]
\centering
\includegraphics[width=\linewidth]{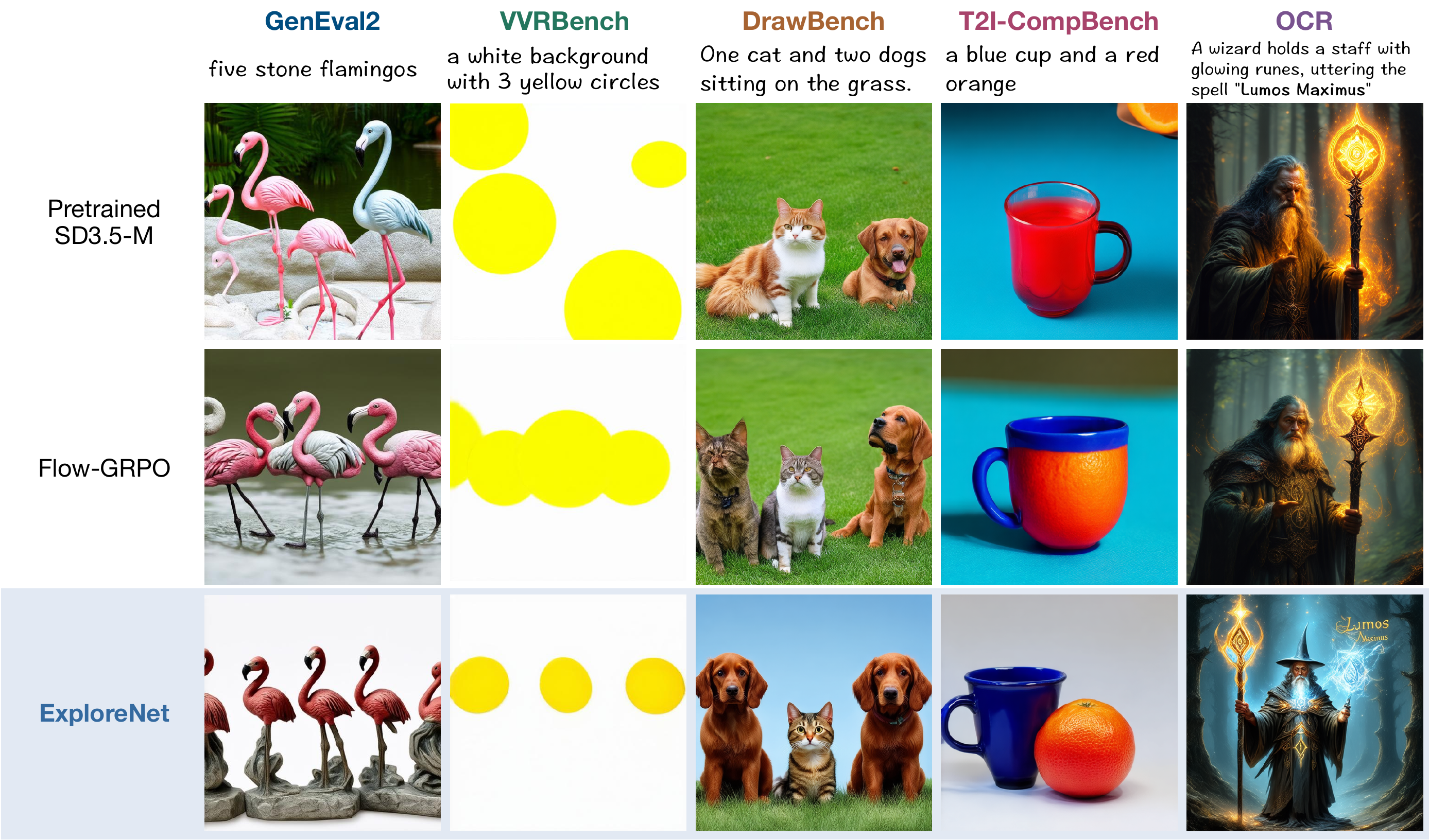}\vspace{-3.5mm}
\caption{\ours satisfies the count, color, text, and material constraints that SD3.5-M and Flow-GRPO miss.
All three annotators preferred \ours to Flow-GRPO on each prompt.}
\label{fig:qualitative-main}\vspace{-5mm}
\end{figure*}

\textbf{The gains transfer beyond the training reward.} \ours is trained only on GenEval2, a multi-constraint composition task scored by visual question answering.
It also improves other benchmarks unseen during training, including GenEval, a template-based composition task scored by an object detector, and VVRBench-Fast, a spatial and topological layout task scored by a model-free verifier.
On VVRBench-Fast, the gain is largest for grounding and spatial relations categories, which check the content and placement of objects (Appendix~\ref{sec:supp-external}).
The gains extend to preference and image-quality models.
Relative to the pretrained model, Flow-GRPO leaves Aesthetic and HPSv2.1 essentially unchanged ($-0.021$ and $-0.001$), whereas \ours raises both ($+0.147$ and $+0.006$).
\ours also improves HPSv3, Aesthetic, and ImageReward on each of four prompt suites.
Consistent gains on the full evaluation suite suggest that \ours improves the generated images themselves rather than exploiting the training reward.

\textbf{Human raters prefer \ours.} Across its comparisons with both baselines, \ours reaches a $67.2\%$ tie-adjusted preference rate, compared with $45.0\%$ for Flow-GRPO and $37.7\%$ for pretrained SD3.5-M (Figure~\ref{fig:human-final-outputs}).
In a separate judgment task, annotators find that \ours changes scene content and composition more than Flow-GRPO relative to matched pretrained out-

\begin{wrapfigure}{r}{0.38\textwidth}\vspace{-4mm}
    \centering
    \includegraphics[width=0.98\linewidth]{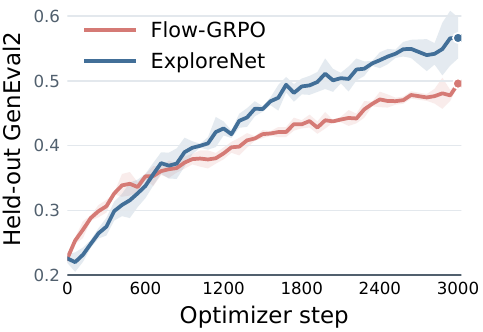}\vspace{-4mm}
    \raggedright
    \captionof{figure}{Held-out GenEval2 during training, avg. over three seeds.
    }
    \label{fig:geneval2-trajectory}\vspace{-7mm}
\end{wrapfigure}

\vspace{-2mm}
puts at a rate of $86.1\%$ (Appendix~\ref{app:displacement}).
Content and composition match
the two most-improved VVRBench-Fast families of grounding and spatial relations.
The favorable human preference and the VVRBench category correlation together indicate that the human-perceived large
divergence of \ours images is toward a positive direction.
Figure~\ref{fig:qualitative-main} shows example generations from the three models over five diverse prompt domains.

\textbf{The gain accumulates during training.} Figure~\ref{fig:geneval2-trajectory} shows how the gain develops.
On held-out GenEval2, \ours trails Flow-GRPO through step 600, leads at every evaluation after, and its lead widens through the end of training.
The gain therefore accumulates over training, consistent with \ours supplying more informative rollout groups at every update, and the widening lead indicates that longer training would not close the gap.
Each \ours update takes $1.25\times$ as long, and at equal wall-clock time \ours still leads: its GenEval2 at update 2400 ($0.532$) exceeds Flow-GRPO's at update 3000 ($0.496$).

\begin{finding}
Exploration in diffusion RL is learnable, and learning it improves the model itself.
\end{finding}

\subsection{Decomposing Magnitude and Shape of Exploration}
\label{sec:results-why}

\ours explores more than Flow-GRPO, with about twice its average noise variance, which follows from the shape of its scale maps (Section~\ref{sec:learned-noise}), and shapes its exploration by allocating noise unevenly across latent elements.
We ask whether the gain comes from the magnitude or the shape of this exploration.
To separate the two, we compare \ours with two alternatives from Section~\ref{sec:setup} that increase exploration without shaping it: the matched isotropic control raises its magnitude, and privileged post-hoc selection raises the number of rollouts.

\textbf{Magnitude without shape lowers performance.} The matched isotropic control adds noise with the same average variance as \ours but spreads it uniformly across latent elements.
On the same training seed, raising the noise uniformly lowers GenEval2 below Flow-GRPO ($0.437$ versus $0.498$), while allocating the same variance raises it to $0.567$ (Table~\ref{tab:controls}).
\ours also exceeds the control on eight of nine metrics.
Two alternative parameterizations let \ours shift the overall level of its noise, one clipping $g_\phi$'s output without standardization and one applying a learned scale and bias after standardization, and both end training below \ours (Appendix~\ref{sec:supp-parameterization}).

\begin{table}[t]\vspace{-3mm}
\caption{Controls and reduced-budget settings.
The main setting uses 24 rollouts per prompt group and 25 denoising steps; the other settings change one of the two.
Each row is a single training run; the main-setting Flow-GRPO and \ours rows are the first training seed of Table~\ref{tab:main-results}, so values differ from the three-seed means there.
}
\label{tab:controls}
\centering
\scriptsize
\setlength{\tabcolsep}{2.6pt}
\resizebox{\textwidth}{!}{%
\begin{tabular}{@{}llrrrrrrrrr@{}}
\toprule
Setting & Method & GenEval2 & VLM Judge & GenEval & VVR-Fast & PickScore & HPSv2.1 & HPSv3 & Aesthetic & ImageReward \\
\midrule
Main & Flow-GRPO & 0.4982 & 4.5500 & 0.6923 & 0.7624 & 0.8447 & 0.3009 & 8.3142 & 5.5024 & 1.1232 \\
Main & Privileged selection & 0.5551 & \textbf{4.7500} & 0.7206 & 0.7724 & 0.8463 & 0.2961 & 8.1998 & 5.4916 & 1.1572 \\
Main & Matched isotropic & 0.4372 & 4.5000 & 0.7351 & 0.7791 & 0.8485 & \textbf{0.3034} & 8.4229 & 5.5104 & 1.1884 \\
Main & \ours & \textbf{0.5674} & 4.7000 & \textbf{0.7452} & \textbf{0.8075} & \textbf{0.8509} & 0.3017 & \textbf{8.6602} & \textbf{5.5997} & \textbf{1.2597} \\
\addlinespace
Group size 8 & Flow-GRPO & 0.4502 & 4.1000 & 0.6771 & 0.7499 & 0.8478 & 0.3006 & 8.2676 & 5.5168 & 1.0916 \\
Group size 8 & \ours & \textbf{0.5389} & \textbf{4.3500} & \textbf{0.7422} & \textbf{0.7952} & \textbf{0.8499} & \textbf{0.3100} & \textbf{8.8990} & \textbf{5.6228} & \textbf{1.2806} \\
\addlinespace
10 steps & Flow-GRPO & 0.4153 & \textbf{3.8875} & 0.6597 & 0.7380 & \textbf{0.8371} & \textbf{0.2804} & 6.9931 & 5.5365 & 0.9521 \\
10 steps & \ours & \textbf{0.4269} & 3.8500 & \textbf{0.7109} & \textbf{0.7562} & 0.8313 & 0.2771 & \textbf{7.1955} & \textbf{5.5765} & \textbf{1.1133} \\
\bottomrule
\end{tabular}%
}
\end{table}

\textbf{\ours matches privileged post-hoc selection with half the rollouts and reward evaluations.} Privileged selection samples 48 rollouts per prompt, observes every reward, and trains on the 24 with the greatest reward spread.
\ours instead sets its noise scales before any reward is observed.
It matches this baseline on GenEval2 ($0.567$ versus $0.555$) and exceeds it on VVRBench-Fast ($0.808$ versus $0.772$), while generating half as many rollouts and requiring half as many reward evaluations; its added cost is one $g_\phi$ evaluation per denoising step.

\begin{finding}
The shape of the exploration distribution outweighs its magnitude.
\end{finding}

\subsection{\ours Remains Effective with Fewer Rollouts and Steps}
\label{sec:results-robustness}

We repeat the comparison with 8 rollouts per prompt instead of 24 and with 10 denoising steps instead of 25 (Table~\ref{tab:controls}).
With 8 rollouts per prompt, \ours leads on every metric, and its GenEval2 ($0.539$) exceeds that of Flow-GRPO with 8 ($0.450$) and with 24 ($0.498$).
With 10 denoising steps, \ours leads on six of nine metrics, including all three compositional benchmarks.

\begin{finding}
Rollout quality is more important than rollout quantity in diffusion RL.
\end{finding}

\subsection{\ours Concentrates Noise on Visually Influential Channels}
\label{sec:structure}

\begin{figure}[t]
\centering
\begin{minipage}[t]{0.48\linewidth}
    \vspace{-4mm}
    \centering
    \includegraphics[width=\linewidth]{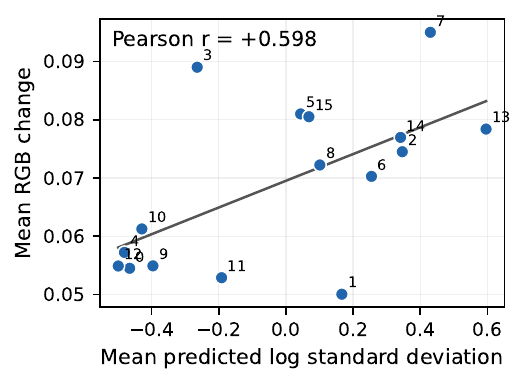}\vspace{-4mm}
    \captionof{figure}{Predicted noise scale tracks channel sensitivity in pixel space.
    Each point averages one channel over 100 prompts.}
    \label{fig:latent-channel-rgb}
\end{minipage}\hfill
\begin{minipage}[t]{0.48\linewidth}
    \vspace{2mm}
    \centering
    \includegraphics[width=\linewidth]{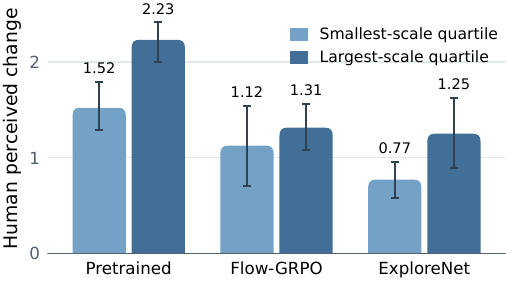}\vspace{-1mm}
    \captionof{figure}{Perturbing \ours's largest-scale channels produces greater human-rated changes than perturbing its smallest-scale channels, in each perturbed denoiser.
    }
    \label{fig:latent-channel-human}
\end{minipage}\vspace{-3mm}
\end{figure}

Section~\ref{sec:results-why} shows that the gain comes from the shape of the exploration distribution, so we analyze the shape of \ours's scale maps at the level of the 16 latent channels, to study how the predicted noise scales relate to each channel's visual effect on the generated image  (Appendix~\ref{sec:supp-latent-channel}).

\textbf{Perturbation setup.} Given a denoiser and a prompt, we generate a reference image with the deterministic ODE sampler. For each latent channel, we resample twice more from the same initial seed while adding (or subtracting) a fixed Gaussian noise to only that channel at every denoising step with all other channels following the ODE update without added noise.
Change is quantified 1) automatically as the mean absolute RGB difference from the reference, and 2) with humans rating how much each perturbed image differs from the reference, from 0 (no visible difference) to 4 (a very large difference).
We compare the change with the noise scale \ours assigns to each channel, obtained by evaluating $g_\phi$ on the latents of the unperturbed reference trajectory and averaging the scale over spatial positions and denoising steps, giving one value per prompt and channel.

\textbf{\ours assigns larger scales to channels that change the image more.}
We perturb every channel of the trained \ours denoiser on 100 GenEval2 training prompts, giving 3,200 perturbed images.
Averaged over prompts, per-channel noise scale and RGB change correlate at $r=+0.60$ across the 16 channels (Figure~\ref{fig:latent-channel-rgb}).
\ours thus concentrates exploration on the channels that most affect the generated image.

\begin{figure}[t]
\centering
\input{figures/sd35_latent_channel_qualitative_grid}\vspace{-3mm}
\caption{Channels to which \ours assigns more noise change the image more.
Perturbing the two channels with the largest \ours noise scales (middle) changes objects, layout, and background.
Perturbing the two with the smallest (right) changes only fine details.
Labels give the average noise multiplier \ours assigns to each channel.
}
\label{fig:latent-channel-qualitative}\vspace{-5mm}
\end{figure}
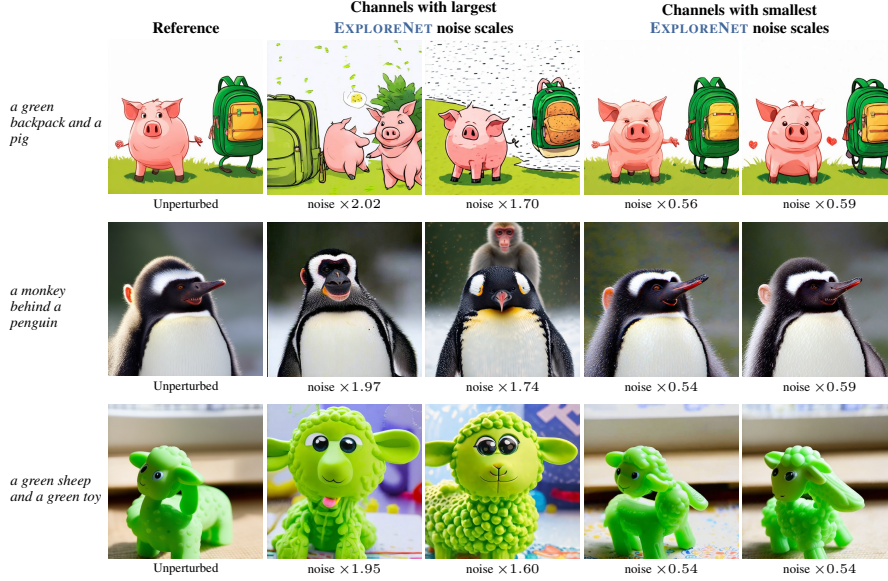

\textbf{Human raters see the same ranking in the pretrained model.}
For each of the pretrained SD3.5-M, Flow-GRPO, and \ours denoisers, we perturb every channel on six prompts, giving 576 perturbed images.
On SD3.5-M images, the mean rating rises from $1.52$ for the channels in the bottom quartile of noise scales to $2.23$ for those in the top quartile, a paired increase of $+0.708$ (Figure~\ref{fig:latent-channel-human}).
Pixel-space change on the pretrained denoiser shows the same pattern at the scale of the main analysis, with a rank correlation between each channel's noise scale and RGB change that is positive in 99 of 100 prompts (mean $\rho=+0.42$).
Channel sensitivity is therefore a property of the SD3.5-M latent space, and \ours's noise scales identify the channels that carry it.
The Flow-GRPO and \ours denoisers show the same ordering (Figure~\ref{fig:latent-channel-human}).
\textbf{The difference is visible in individual images.} Figure~\ref{fig:latent-channel-qualitative} shows interventions on the two largest-scale and two smallest-scale channels of each prompt.
Perturbing the largest-scale channels produces visibly larger changes.

\begin{finding}
Generative latent spaces are unevenly sensitive, and exploration should follow that structure.
\end{finding}

%% file: figures/sd35_latent_channel_qualitative_grid.tex
\begingroup
\setlength{\tabcolsep}{1pt}
\renewcommand{\arraystretch}{0.92}
\newcommand{\latentprompt}[1]{%
  \parbox[c]{0.105\textwidth}{\raggedright\fontsize{6.5}{7.2}\selectfont\textit{#1}}}
\newcommand{\latentimage}[2]{%
  \begin{minipage}[c]{0.172\textwidth}
    \centering
    \includegraphics[width=\linewidth]{figures/latent_interventions/#1}\par
    \vspace{0pt}
    {\fontsize{5.7}{6.3}\selectfont #2\par}
  \end{minipage}}
\resizebox{0.85\linewidth}{!}{
\begin{tabular}{@{}lccccc@{}}
& {\fontsize{7}{8}\selectfont\textbf{Reference}}
& \multicolumn{2}{c}{\fontsize{7}{8}\selectfont\shortstack{\textbf{Channels with largest}\\\textbf{\ours noise scales}}}
& \multicolumn{2}{c}{\fontsize{7}{8}\selectfont\shortstack{\textbf{Channels with smallest}\\\textbf{\ours noise scales}}} \\
\latentprompt{a green backpack and a pig}
& \latentimage{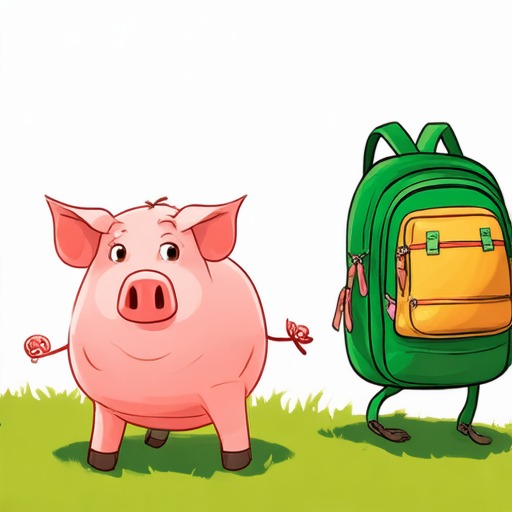}{Unperturbed}
& \latentimage{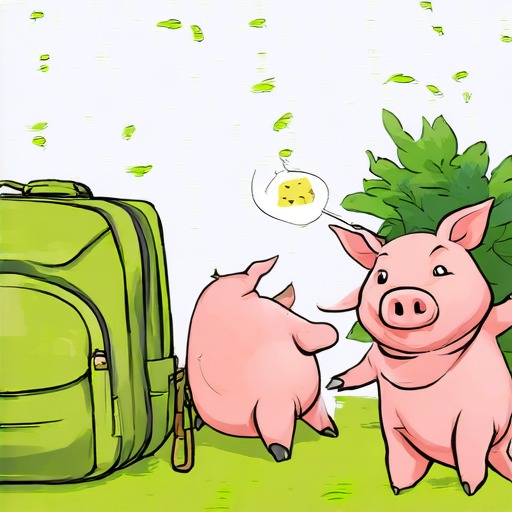}{noise $\times2.02$}
& \latentimage{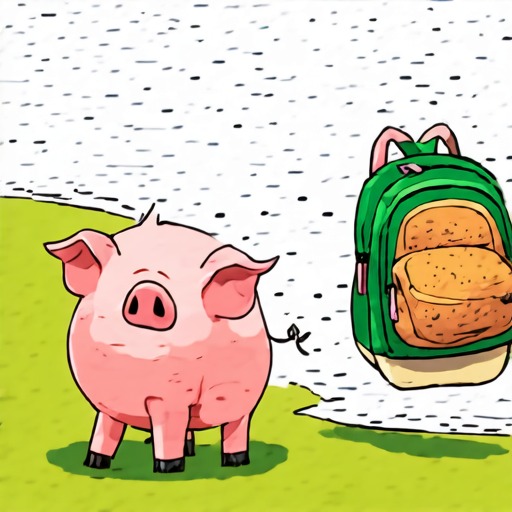}{noise $\times1.70$}
& \latentimage{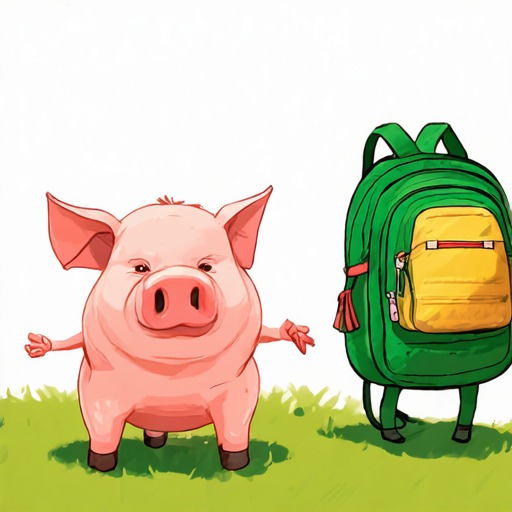}{noise $\times0.56$}
& \latentimage{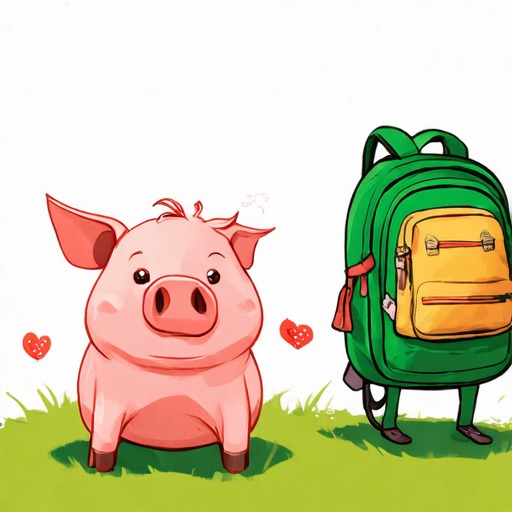}{noise $\times0.59$} \\ \noalign{\vspace{5pt}}
\latentprompt{a monkey behind a penguin}
& \latentimage{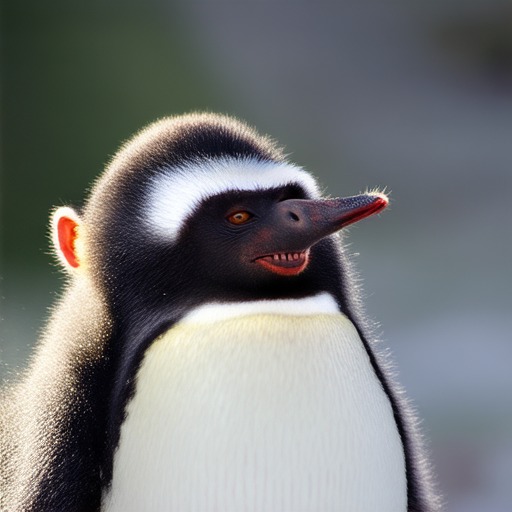}{Unperturbed}
& \latentimage{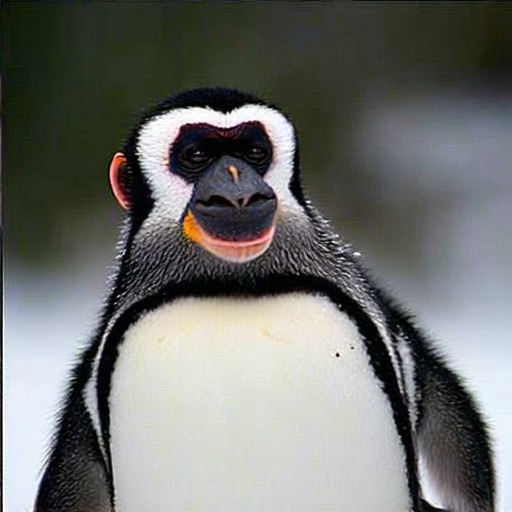}{noise $\times1.97$}
& \latentimage{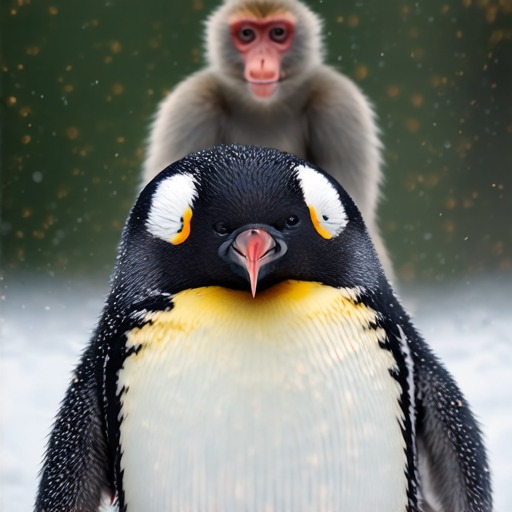}{noise $\times1.74$}
& \latentimage{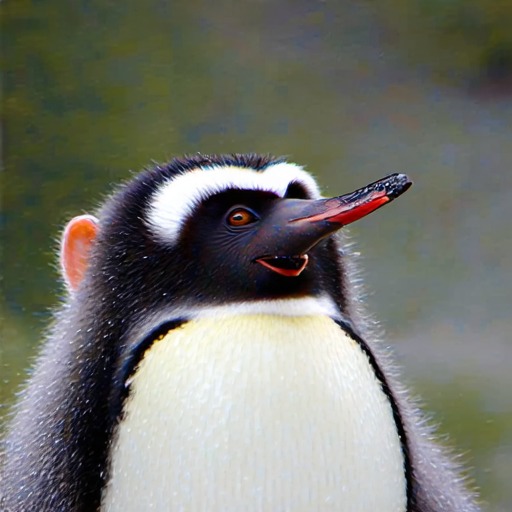}{noise $\times0.54$}
& \latentimage{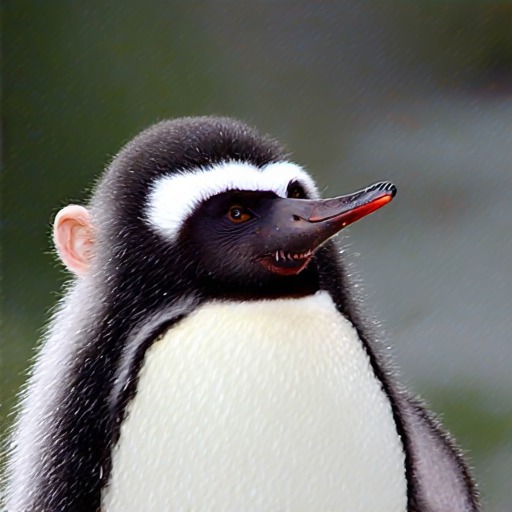}{noise $\times0.59$} \\ \noalign{\vspace{5pt}}
\latentprompt{a green sheep and a green toy}
& \latentimage{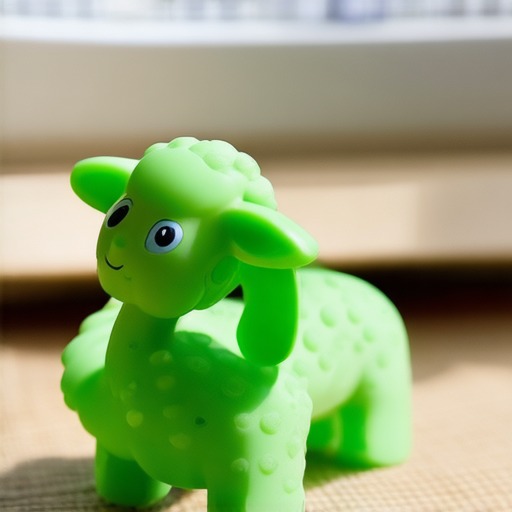}{Unperturbed}
& \latentimage{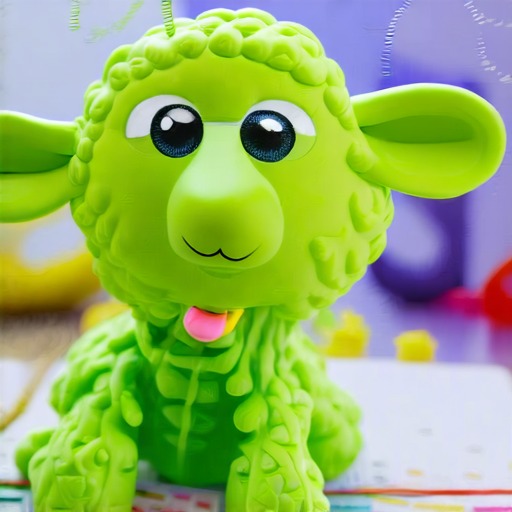}{noise $\times1.95$}
& \latentimage{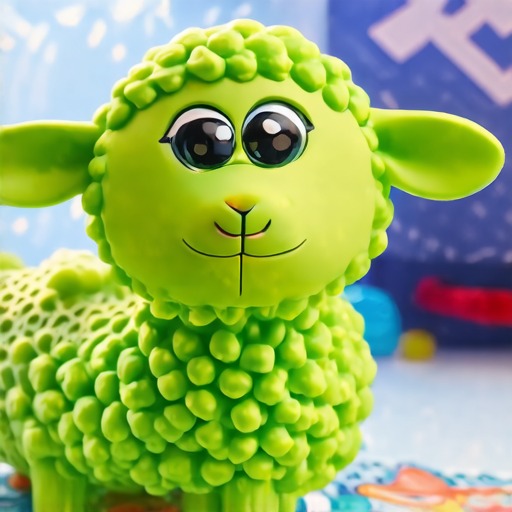}{noise $\times1.60$}
& \latentimage{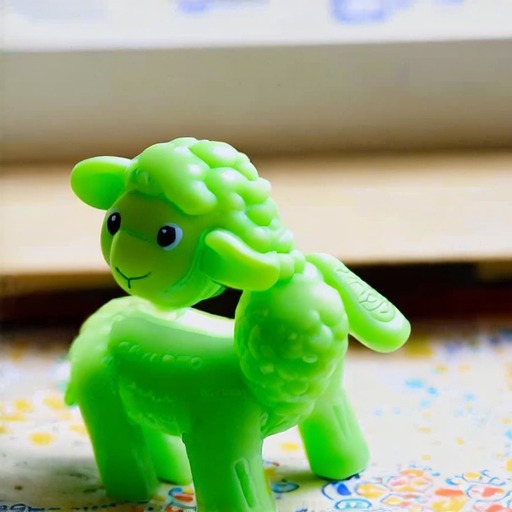}{noise $\times0.54$}
& \latentimage{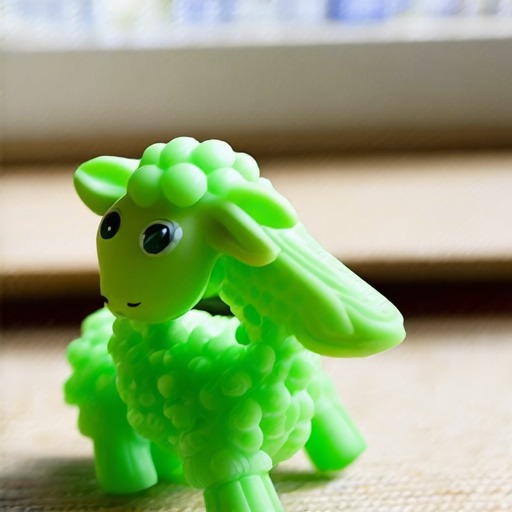}{noise $\times0.54$}
\end{tabular}
}
\endgroup

%% file: sec/1_related_work.tex
\section{Related Work}
\label{sec:related}\vspace{-1mm}

\vspace{-0.5mm}
\textbf{Reward-based post-training of diffusion and flow models.} Diffusion and flow models have been post-trained with reward-weighted likelihood \citep{lee2023humanfeedback}, policy gradients \citep{black2023ddpo,fan2023dpok,zhang2024largescalerl}, differentiable rewards \citep{clark2023draft,prabhudesai2023alignprop}, preference optimization \citep{wallace2023diffusiondpo,yang2023d3po,yang2024densereward}, and KL- or entropy-regularized stochastic control \citep{uehara2024entropycontrol,domingoenrich2024adjoint}.
GRPO \citep{shao2024deepseekmath} was adapted to flow models by Flow-GRPO and DanceGRPO, which convert the deterministic sampler into a stochastic one \citep{liu2025flowgrpo,xue2025dancegrpo}, and later methods learn the reward adversarially or optimize group preferences directly \citep{mao2026image,luo2026reinforcing}.
Follow-up work changes which denoising steps explore, through sliding stochastic windows, branching, or high-entropy steps \citep{li2025mixgrpo,li2025branchgrpo,zhang2026egrpo}, how much each step explores, as DenseGRPO does by calibrating a single noise level per step in a separate pass over sampled trajectories \citep{deng2026densegrpo,zheng2026sagegrpo}, or how rewards are credited to steps \citep{he2025tempflowgrpo,tong2026turningpoint,savani2026stepwiseflow,yan2026sgpo,deng2026densegrpo}.
\ours instead learns the shape of the exploration distribution across latent elements from the current state; its scale map multiplies the per-step noise coefficient $a_t$, so a per-step schedule sets the level of the noise at each step and \ours sets its allocation within the step.

\vspace{-0.5mm}
\textbf{Constructing informative rollouts.} Post-hoc selection methods observe the rewards of additional rollouts, with ProGRPO keeping subsets of high reward spread and Smart-GRPO searching over candidate noise \citep{ge2025progrpo,yu2025smartgrpo}.
DiffExp perturbs classifier-free guidance and prompt-phrase weights, and E$^2$PO optimizes prompt embeddings for semantic spread \citep{chae2025diffexp,hu2026e2po}.
Initial-noise methods optimize the starting latent of a sample or learn prompt-conditioned initial-noise distributions \citep{guo2024initno,eyring2024reno,chen2024find,miao2025noiseppo, eyring2025noisehypernet}.
These approaches spend additional rollouts and reward evaluations, perturb the conditioning, or act before denoising begins.
\ours sets the noise at every denoising step, before any reward is observed, and is trained jointly with the denoiser.

\vspace{-0.5mm}
\textbf{Learned noise and exploration.} Diffusion models have learned reverse variances and covariances \citep{nichol2021improved,ou2025ocm}, data-dependent forward processes \citep{sahoo2024mulan}, and spatially adaptive or reweighted noise \citep{lavda2026sani,wu2026noiserater}, and inference-time samplers add flow-orthogonal perturbations or learn noise schedules \citep{wu2025trajectoryspread,lim2026lesamp}; these methods shape noise to model the data or to improve samples.
In flow-model RL, Coefficients-Preserving Sampling, related analyses, and Precise study how much stochasticity the sampler should inject, with the noise fixed by design \citep{wang2025cps,sheng2025samplerstochasticity,zou2026precise}, and diversity-preserving objectives regularize what the model is trained toward \citep{liu2025gradientgfn,sorokin2025imagerefl,liu2025diversegrpo,liu2026drift,chen2025d2align}.
In deep RL, parameter noise and state-dependent exploration replace fixed action noise with structured perturbations \citep{plappert2017paramnoise,fortunato2017noisynet,raffin2020gsde}.
\ours brings state-dependent exploration to diffusion RL, learning the exploration distribution from the reward spread of each rollout group and discarding it at inference.

%% file: sec/5_conclusion.tex
\vspace{-1mm}
\section{Conclusion}
\label{sec:conclusion}\vspace{-1mm}

Group-relative diffusion RL learns only from the alternatives its sampler produces, yet the exploration distribution that produces them follows a fixed schedule that treats every latent element alike.
Latent elements differ in how much they change the generated image, and \ours makes the exploration distribution trainable so that it can adapt to these differences.
It predicts a noise scale for every latent element from the current state of generation, learns from the reward spread of each rollout group jointly with the denoiser, and is discarded after training, so the final model is sampled exactly as before.
Across our experiments, exploration proves learnable, the shape of the exploration distribution outweighs its magnitude, and rollout quality is more effective than rollout quantity.
Exploration is therefore a design axis of diffusion RL in its own right, alongside reward design and per-step noise schedules.

\section*{Limitations}

The exploration distribution $q_\phi$ is a diagonal Gaussian; correlated noise across latent elements, such as low-rank or spatially structured covariances, could explore coherent changes such as moving an object as a whole.
\ours learns only the transition noise, and the same framework could learn other sources of rollout variation, such as the guidance scale or the initial noise, jointly with it.

\ours learns from the reward spread of each rollout group.
Other group statistics, such as the spread of individual reward components or of verifier constraint families, could train separate exploration distributions for different objectives.

Video latents add a temporal axis, and \ours can allocate exploration across frames as well as channels and spatial positions.
Group-relative RL for autoregressive and discrete-diffusion generators explores by sampling tokens at a fixed temperature, and a learned per-position temperature could play the role of \ours's scale map.

\ours is a second SD3.5-M transformer; a lightweight head on the denoiser could predict the same scale maps from its own features.
Channel sensitivity is a property of the SD3.5-M latent space, so a trained \ours could be reused across rewards and fine-tuning runs of the same base model.
The scale maps identify which latent channels control the content of a generated image, which could guide diverse sampling, localized editing, and the perturbations of best-of-$N$ and other inference-time search methods.

A theoretical account of when the shape of the exploration distribution outweighs its magnitude, relating latent sensitivity to the reward spread of rollout groups, remains open.

\section*{Ethics Statement}
\ours improves reward-based fine-tuning of text-to-image models and inherits the risks of the models it fine-tunes, including the generation of misleading or harmful images.
The human evaluation asked annotators to compare generated images and to rate visual changes; the annotators were graduate students who volunteered, and Appendix~\ref{sec:supp-human-evaluation} describes the protocol.

\section*{Reproducibility Statement}
Section~\ref{sec:learned-noise} and Appendix~\ref{sec:supp-training-details} specify the objectives, architecture, and all training hyperparameters.
Appendix~\ref{sec:supp-evaluation-details} specifies the prompt sets, metric implementations, generation settings, and aggregation.
All code is released.

\section*{Acknowledgment}
This research was developed in part with funding from the Defense Advanced Research Projects Agency's (DARPA) SciFy program (Agreement No. HR00112520300). The views expressed are those of the author and do not reflect the official policy or position of the Department of Defense or the U.S.~Government.
This material is based in part upon work supported by the Defense Advanced Research Projects Agency and the Air Force Research Laboratory, contract number(s): FA8650-23-C-7316. Any opinions, findings and conclusions, or recommendations expressed in this material are those of the author(s) and do not necessarily reflect the views of AFRL or DARPA.
This research was supported by Coefficient Giving, the University of Washington Population Health Initiative, Amazon Health, the UW+Amazon Science Hub, and the Meta AIM program.

%% file: sec/X_suppl.tex
\section{Training Details}
\label{sec:supp-implementation}

\subsection{Hyperparameters and Compute}
\label{sec:supp-training-details}

The image experiments use SD3.5-M at $512\times512$ resolution with 25 sampling and evaluation denoising steps, classifier free guidance $4.5$, Flow-GRPO noise level $\lambda=0.7$, and 24 rollouts per prompt in the main setting.
Training uses 64 H100 GPUs over eight nodes.
Each rank samples microbatches of 3, with 8 sampling batches per epoch, one inner epoch, and gradient accumulation 4.
In the main setting, each training iteration therefore samples $P=64$ prompts with $G=24$ rollouts each (1,536 images), and both policies take two optimizer updates on these rollouts. The update counts reported throughout the paper are optimizer updates.
The group-size-8 runs keep this schedule and 1,536 images per iteration, so each iteration samples $P=192$ prompts with $G=8$ rollouts each.
Both the denoiser and \ours use rank 32 LoRA adapters with LoRA scale 64 on the SD3.5-M attention projections.
The denoising optimizer uses Adam with learning rate $3\times10^{-4}$, $\beta_1=0.9$, $\beta_2=0.999$, weight decay $10^{-4}$, and gradient norm clipping at 1.0.

\ours initializes a separate SD3.5-M transformer from the pretrained model, freezes its base weights, and trains its rank 32 LoRA parameters in fp32.
Its Adam optimizer uses learning rate $10^{-4}$ and zero weight decay.
The final parameterization normalizes the output per sample and clips log standard deviation directly to $[-1,1]$, with a standard-deviation floor of $10^{-6}$ during normalization.
Reward normalization uses an additive constant of $10^{-4}$.
Both clipped objectives use clip range $\epsilon=10^{-4}$ and advantages clipped to $[-5,5]$.
The exploration objective applies a KL penalty of weight $\beta_\phi=10^{-3}$ toward the unit Gaussian, and the denoising objective applies a KL penalty of weight $\beta_\theta=0.04$ toward the pretrained denoiser, for both Flow-GRPO and \ours.
The denoising and exploration policy optimizers are stepped after the same backward pass.

The 48-to-24 privileged post-hoc selection baseline doubles rollout generation and GenEval2 reward calls relative to the 24 rollout runs, then keeps the 12 lowest- and 12 highest-reward samples, the subset of 24 with the greatest reward spread.

Measured over steps 2160 through 2580, Flow-GRPO requires $151.7$ seconds per optimizer update and \ours requires $189.4$ seconds.
\ours therefore takes $1.25\times$ as long per update, corresponding to $161.8$ versus $202.0$ GPU hours per 60 updates, including periodic evaluation and checkpointing.
During training, \ours holds a second 2.5-billion-parameter SD3.5-M transformer in memory, and both methods fit a per-rank microbatch of 3 on 80 GB H100 GPUs.
At inference, both methods use 25 denoiser evaluations and \ours is not used.

\begin{table}[H]
\centering
\small
\setlength{\tabcolsep}{7pt}
\begin{tabular}{@{}lrrrr@{}}
\toprule
Setting & Rollouts & Reward calls & \shortstack{Denoising\\steps} & \shortstack{\ours\\evaluations} \\
\midrule
Main Flow-GRPO training & 24 & 24 & 600 & 0 \\
Main \ours training & 24 & 24 & 600 & 600 \\
Privileged selection 48 to 24 training & 48 & 48 & 1200 & 0 \\
Matched isotropic training & 24 & 24 & 600 & 0 \\
Group size 8 training & 8 & 8 & 200 & 200 \\
10 step training & 24 & 24 & 240 & 240 \\
Flow-GRPO inference & 1 & 0 & 25 & 0 \\
\ours inference & 1 & 0 & 25 & 0 \\
\bottomrule
\end{tabular}
\caption{\ours adds $g_\phi$ evaluations during training while both trained denoisers retain the same inference cost.
Counts describe the sampling passes for one training prompt group or one inference image.}
\label{tab:supp-compute-budget}
\end{table}

\subsection{Noise-Scale Parameterization}
\label{sec:supp-parameterization}

This ablation tests how the output of $g_\phi$, $h_\phi$, is converted into the elementwise log standard deviation $\ell_\phi$.
The \emph{normalized affine} variant standardizes $h_\phi$ as in Eq.~\ref{eq:explorenet-normalize}, applies a learned scalar scale and bias, and clips the result to $[-1,1]$.
The \emph{raw clipped} variant uses the transformer output directly and clips it to the same range.
The final \ours parameterization standardizes the output and clips it directly, without the learned scalar scale and bias, as in Eq.~\ref{eq:explorenet-clamp}.
In each case, exponentiating $\ell_\phi$ gives the standard deviation of the zero-mean Gaussian transition noise.

All four rows use the same prompts, 25-step sampler, rollout group size, and microbatch-3 optimization schedule.
Table~\ref{tab:supp-parameterization} reports one 80-prompt GenEval2 evaluation at an intermediate checkpoint and at step 3060; each row is a single training run.
We use normalized-and-clipped noise in the main experiments because it performs best at both checkpoints in this controlled comparison.

\begin{table}[H]
\centering
\small
\setlength{\tabcolsep}{5pt}
\begin{tabular}{@{}lrr@{}}
\toprule
Noise parameterization & Step 1140 & Step 3060 \\
\midrule
Flow-GRPO (isotropic) & 0.3825 & 0.5146 \\
\ours, normalized affine & 0.4009 & 0.5057 \\
\ours, raw clipped & 0.4080 & 0.5358 \\
\ours, normalized and clipped & \textbf{0.4179} & \textbf{0.5679} \\
\bottomrule
\end{tabular}
\caption{GenEval2 scores for Flow-GRPO and three mappings from the output of $g_\phi$ to log standard deviation.}
\label{tab:supp-parameterization}
\end{table}

\subsection{Reward-Spread Selection}
\label{sec:supp-spread-selection}

The selection experiment in Section~\ref{sec:learned-noise} draws 16 rollouts per prompt and trains on 8, chosen by smallest reward spread, at random, or by greatest reward spread.
Greatest-spread selection keeps the four lowest- and four highest-reward rollouts, which maximizes the reward spread of the kept rollouts; smallest-spread selection keeps the eight rollouts with consecutive rewards of lowest variance.
The three conditions share the rollout pool, group size, optimizer schedule, and training length, with one training run each; this schedule is separate from the main experiments, so its values compare the selection rules with one another.
Table~\ref{tab:spread-selection} evaluates the three final models on the full benchmark suite with fixed prompts and generation seed.

\begin{table}[H]
\centering
\small
\setlength{\tabcolsep}{6pt}
\begin{tabular}{@{}lrrr@{}}
\toprule
Metric & Smallest & Random & Greatest \\
\midrule
GenEval2 & 0.2579 & 0.3859 & \textbf{0.4648} \\
VLM Judge & 3.8750 & 4.4625 & \textbf{4.5125} \\
GenEval & 0.6171 & 0.6537 & \textbf{0.6948} \\
VVR-Fast & 0.7210 & 0.7380 & \textbf{0.7614} \\
PickScore & 0.8414 & 0.8433 & \textbf{0.8456} \\
HPSv2.1 & 0.3016 & 0.3054 & \textbf{0.3073} \\
HPSv3 & 7.8131 & 8.1597 & \textbf{8.3543} \\
Aesthetic & 5.5201 & \textbf{5.5228} & 5.5122 \\
ImageReward & 0.9409 & 1.0517 & \textbf{1.1075} \\
\bottomrule
\end{tabular}
\caption{Benchmark results of the three selection rules.
Greatest-spread selection is best on eight of nine metrics.}
\label{tab:spread-selection}
\end{table}

\subsection{Matched Isotropic Control}
\label{sec:supp-varmatch}

The control samples each transition with Eq.~\ref{eq:explorenet-transition}, setting $s_j=1.410490688$ for every latent element, so its noise covariance is $1.989483981\,a_t^2 I$.
This matches the mean variance predicted across latent elements and the three main \ours seeds over training steps 2760--2999.
As in \ours, the transition mean is Flow-GRPO's $\mu_\theta$.
The control has no $g_\phi$ and uses the seed 1 training schedule through step 3000.

\section{Evaluation Protocol}
\label{sec:supp-evaluation}

\subsection{Automatic Evaluation}
\label{sec:supp-evaluation-details}

\paragraph{Benchmarks and prompt sets.} GenEval2~\citep{kamath2025geneval2} contains 720 training prompts and a fixed 80-prompt held-out split covering object presence, count, attributes, and spatial relations.
Each prompt is decomposed into atomic visual questions.
Qwen3-VL-2B assigns probability to the accepted short answers, and the image score is the geometric mean across questions.
GenEval2 is the sole reward used to train the main models.
On the same 80-prompt test set, VLM Judge scores prompt alignment from 1 to 5.

Official GenEval~\citep{ghosh2023geneval} uses 553 prompts with four generations per prompt, giving 2,212 evaluated images.
VVRBench-Fast~\citep{verifiablevisualrewards2026} contains 820 prompts sampled from VVRBench Core across its attainable complexity levels.
Its frozen, model-free verifier evaluates grounding, cardinality, spatial, size, and topology constraints from executable scene specifications.
We report dense VVR, the mean dense verifier score over all 820 prompts.

We evaluate PickScore~\citep{kirstain2023pickscore} on 500 unique prompts from the Pick-a-Pic v1 validation split.
We evaluate HPSv2.1~\citep{wu2023human} on its complete 3,200-prompt benchmark, comprising 800 prompts each for anime, concept art, paintings, and photo.
We report the mean PickScore, the four HPS domain means, and their unweighted macro average.

We evaluate HPSv3~\citep{ma2025hpsv3}, Aesthetic score, and ImageReward on the full 200-prompt DrawBench~\citep{saharia2022imagen} set and fixed 1,000-prompt lists from PartiPrompts~\citep{yu2022parti}, DPG-Bench~\citep{hu2024ella}, and T2I-CompBench~\citep{huang2023t2icompbench}.
The aesthetic score uses the LAION improved aesthetic predictor~\citep{schuhmann2022aesthetic}, with a CLIP ViT-L/14 image encoder and its released linear MSE head; ImageReward uses ImageReward-v1.0~\citep{xu2023imagereward}.
We average each metric within a prompt list and report the unweighted mean of the four list means.

\paragraph{Generation and aggregation.} Checkpoints are saved and evaluated every 60 optimizer updates.
For the main comparison, we evaluate the checkpoint at step 3000 without selecting by evaluation score.
Evaluation uses the trained denoiser with the deterministic flow-matching sampler without \ours.
The primary comparison comprises three independent training seeds per method and three GenEval2 evaluation passes per seed.
All other automatic benchmarks use one pass per training seed.
Each model generates one image per prompt except for official GenEval, which uses four.
Matched methods receive the same prompts and initial latent seeds.
We average evaluation passes within each training seed, then report the mean and sample standard deviation across seeds.
Controlled and robustness experiments use one training seed unless their corresponding result states otherwise.
Flow-GRPO and \ours runs are paired by training seed (42, 123, and 456, labeled 1 to 3); Table~\ref{tab:supp-per-seed} reports every seed.

\subsection{Human Evaluation}
\label{sec:supp-human-evaluation}

The preference and displacement studies use the same 72 matched triplets of pretrained, Flow-GRPO, and \ours outputs described in Section~\ref{sec:experiments}.
Each of the six prompt suites contributes 12 examples, and each training seed contributes four examples within a suite.
For DrawBench, PartiPrompts, DPG-Bench, T2I-CompBench, and the Flow-GRPO OCR prompt set~\citep{liu2025flowgrpo}, we draw 12 distinct prompts uniformly without replacement using a fixed study seed.
For VVRBench, we draw one prompt from each of 12 prespecified strata spanning composition tier and difficulty.
We then shuffle each suite with the same fixed seed and assign prompts cyclically across the three training seeds.
The panel is fixed before image generation, and the three models use the same generation seed for each prompt.

The preference interface presents the prompt and two unlabeled candidates.
It instructs annotators to prioritize satisfaction of the full prompt and then visual quality, reserving a tie for cases in which neither image is meaningfully preferable.
Trial order and candidate position are independently randomized.
Three annotators evaluate all 216 pairwise comparisons, giving 648 judgments.
Preference rates count each tie as one half for each model.
Confidence intervals resample prompts while retaining all ratings associated with each sampled prompt.

Pairwise exact agreement among the three annotators ranges from $56.9\%$ to $72.7\%$, and $47.7\%$ of comparisons receive unanimous labels.

\begin{table}[H]
\centering
\small
\setlength{\tabcolsep}{3pt}
\begin{tabular}{@{}lrrrrr@{}}
\toprule
Preferred model & Wins & Ties & Losses & Rate & 95\% CI \\
\midrule
Flow-GRPO over pretrained & 118 & 8 & 90 & $56.5\%$ & $[48.6,64.4]$ \\
\ours over pretrained & 143 & 8 & 65 & $68.1\%$ & $[60.0,75.9]$ \\
\ours over Flow-GRPO & 140 & 7 & 69 & $66.4\%$ & $[58.6,74.1]$ \\
\bottomrule
\end{tabular}
\caption{\ours wins each pairwise preference comparison.
Each row contains 216 ratings over 72 prompts.
Rates split ties equally, and confidence intervals resample prompts with all associated ratings.}
\label{tab:supp-human-preference}
\end{table}

\paragraph{Qualitative examples.}
Figure~\ref{fig:qualitative-main} shows three prompts from the preference study (DrawBench, T2I-CompBench, and OCR), on which all three annotators preferred \ours to Flow-GRPO and each model's image uses the shared generation seed, and two prompts from the training-time evaluations (VVRBench and held-out GenEval2), whose images three annotators ranked separately, each preferring \ours.
Its OCR header abbreviates the prompt ``A wizard holds a staff with glowing runes, casting a bright light as he utters the spell `Lumos Maximus', illuminating a dark, mystical forest.''

\paragraph{Change from pretrained outputs.}
\label{app:displacement}
The displacement study uses the same 72 matched triplets and three complete annotators.
The interface presents the pretrained output as a labeled reference between blinded Flow-GRPO and \ours candidates.
Annotators identify which candidate differs more from the reference in scene content or composition.
The instructions define this criterion through entities, attributes, counts, spatial arrangement, viewpoint, and framing, while excluding differences limited to color, lighting, texture, rendering style, or image quality.
Trial order and candidate position are independently randomized for each annotator.
Across 216 ratings, \ours is selected 181 times, Flow-GRPO 25 times, and 10 judgments are ties.
The tie-adjusted \ours rate is $86.1\%$, with a 95\% CI of $[80.1,91.4]$.
\ours receives a majority on 62 of 72 prompts, Flow-GRPO on 6, and 4 prompts have no majority.
Pairwise exact agreement ranges from $77.8\%$ to $80.6\%$, with unanimous labels on $70.8\%$ of prompts.

\section{Additional Results}
\label{sec:supp-results}

\subsection{Per-Seed Results}
\label{sec:supp-per-seed}

\begin{table}[H]
\centering
\scriptsize
\setlength{\tabcolsep}{2.6pt}
\resizebox{\linewidth}{!}{%
\begin{tabular}{@{}llrrrrrrrrr@{}}
\toprule
Method & Seed & GenEval2 & \shortstack{VLM\\Judge} & GenEval & VVR-Fast & PickScore & HPSv2.1 & HPSv3 & Aesthetic & ImageReward \\
\midrule
Flow-GRPO & 1 & 0.4982 & 4.5500 & 0.6923 & 0.7624 & 0.8447 & 0.3009 & 8.3142 & 5.5024 & 1.1232 \\
Flow-GRPO & 2 & 0.4933 & 4.5750 & 0.7062 & 0.7715 & 0.8467 & 0.3004 & 8.3142 & 5.5184 & 1.1339 \\
Flow-GRPO & 3 & 0.4956 & 4.4000 & 0.6879 & 0.7598 & 0.8430 & 0.2966 & 8.0051 & 5.4673 & 1.0840 \\
\addlinespace
\ours & 1 & 0.5674 & 4.7000 & 0.7452 & 0.8075 & 0.8509 & 0.3017 & 8.6602 & 5.5997 & 1.2597 \\
\ours & 2 & 0.5972 & 4.7750 & 0.7335 & 0.7928 & 0.8509 & 0.3056 & 8.7869 & 5.6744 & 1.2841 \\
\ours & 3 & 0.5337 & 4.7125 & 0.7261 & 0.7703 & 0.8519 & 0.3097 & 8.9658 & 5.7175 & 1.2883 \\
\bottomrule
\end{tabular}%
} \caption{Per-seed results at step 3000 underlying Table~\ref{tab:main-results}.
GenEval2 averages three evaluation passes; other metrics use one pass.}
\label{tab:supp-per-seed}
\end{table}

\subsection{Late-Training Stability and Prompt-Level Uncertainty}
\label{sec:supp-prompt-bootstrap}

We measure stability over the final five checkpoints by averaging steps 2760, 2820, 2880, 2940, and 3000 within each evaluation pass and training seed.
The result then averages three evaluation passes within each seed and reports the mean and sample standard deviation across the three training seeds.

\begin{table}[H]
\centering
\small
\setlength{\tabcolsep}{7pt}
\begin{tabular}{@{}lrrr@{}}
\toprule
Metric & Flow-GRPO & \ours & Gain \\
\midrule
GenEval2 & $0.4809_{\pm0.0093}$ & $\mathbf{0.5524_{\pm0.0300}}$ & $+0.0715$ \\
\bottomrule
\end{tabular}
\caption{The GenEval2 gain persists across the final five checkpoints.
Each seed value averages checkpoints 2760 through 3000 over three evaluation passes.
Subscripts report sample standard deviation across training seeds.}
\label{tab:supp-late-training}
\end{table}

We also estimate sensitivity to the composition of each evaluation set with paired prompt bootstraps.
For GenEval2, two evaluation passes retained complete prompt-level records for all three training seeds, both methods, and checkpoints 2760 through 3000; each replicate preserves these pairings and averages over training seeds, checkpoints, and evaluation passes.
For VVRBench-Fast, each replicate resamples the 820 matched checkpoint-3000 prompts and averages the paired difference across the three training-seed pairs.
Both analyses use 200,000 replicates and condition on the six trained models.

\begin{table}[H]
\centering
\small
\setlength{\tabcolsep}{5pt}
\begin{tabular}{@{}lrr@{}}
\toprule
Evaluation & Paired gain & 95\% prompt bootstrap interval \\
\midrule
GenEval2 & $+0.0748$ & $[+0.0405,+0.1085]$ \\
VVR-Fast & $+0.0256$ & $[+0.0202,+0.0311]$ \\
\bottomrule
\end{tabular}
\caption{Prompt resampling gives positive gain intervals for GenEval2 and VVRBench-Fast.}
\label{tab:supp-prompt-bootstrap}
\end{table}

\subsection{Per-Benchmark Results}
\label{sec:supp-external}

Table~\ref{tab:supp-standard-preference} reports PickScore and the four HPSv2.1 styles on their benchmark prompts, and Table~\ref{tab:supp-external-quality} reports HPSv3, Aesthetic, and ImageReward on each of the four prompt suites.
Each entry is the mean and sample standard deviation across the three training seeds, and each seed generates one image per prompt with evaluation seed 42.

\begin{table}[H]
\centering
\small
\setlength{\tabcolsep}{6pt}
\begin{tabular}{@{}lrrr@{}}
\toprule
Prompt set and metric & Flow-GRPO & \ours & Gain \\
\midrule
Pick-a-Pic v1 (500), PickScore
& $0.8448_{\pm0.0019}$ & $\mathbf{0.8512_{\pm0.0006}}$ & $+0.0064$ \\
HPSv2 anime prompts (800)
& $0.3130_{\pm0.0018}$ & $\mathbf{0.3204_{\pm0.0038}}$ & $+0.0073$ \\
HPSv2 concept art prompts (800)
& $0.3034_{\pm0.0026}$ & $0.3092_{\pm0.0054}$ & $+0.0058$ \\
HPSv2 paintings prompts (800)
& $0.3038_{\pm0.0027}$ & $0.3085_{\pm0.0039}$ & $+0.0047$ \\
HPSv2 photo prompts (800)
& $0.2770_{\pm0.0027}$ & $\mathbf{0.2845_{\pm0.0032}}$ & $+0.0075$ \\
HPSv2.1 macro (3,200)
& $0.2993_{\pm0.0023}$ & $0.3056_{\pm0.0040}$ & $+0.0063$ \\
\bottomrule
\end{tabular}
\caption{\ours scores higher on PickScore and on all four HPSv2.1 prompt styles.
Bold marks gains whose displayed intervals do not overlap.}
\label{tab:supp-standard-preference}
\end{table}

\begin{table}[H]
\centering
\small
\setlength{\tabcolsep}{5pt}
\begin{tabular}{@{}llrrr@{}}
\toprule
Prompt suite & Method & HPSv3 & Aesthetic & ImageReward \\
\midrule
DrawBench
& Flow-GRPO & $9.5718_{\pm0.2295}$ & $5.3664_{\pm0.0279}$ & $1.0602_{\pm0.0503}$ \\
& \ours & $\mathbf{10.3331_{\pm0.2491}}$ & $\mathbf{5.5109_{\pm0.0625}}$ & $\mathbf{1.2683_{\pm0.0210}}$ \\
\addlinespace
PartiPrompts
& Flow-GRPO & $7.4324_{\pm0.1504}$ & $5.5783_{\pm0.0275}$ & $1.2560_{\pm0.0110}$ \\
& \ours & $\mathbf{7.8012_{\pm0.0808}}$ & $\mathbf{5.7438_{\pm0.0507}}$ & $\mathbf{1.3872_{\pm0.0174}}$ \\
\addlinespace
DPG-Bench
& Flow-GRPO & $9.2953_{\pm0.2166}$ & $5.6607_{\pm0.0287}$ & $0.8724_{\pm0.0316}$ \\
& \ours & $\mathbf{9.8481_{\pm0.0996}}$ & $\mathbf{5.8578_{\pm0.0401}}$ & $\mathbf{1.0256_{\pm0.0153}}$ \\
\addlinespace
T2I-CompBench
& Flow-GRPO & $6.5454_{\pm0.1530}$ & $5.3788_{\pm0.0253}$ & $1.2663_{\pm0.0285}$ \\
& \ours & $\mathbf{7.2348_{\pm0.2011}}$ & $\mathbf{5.5428_{\pm0.0885}}$ & $\mathbf{1.4283_{\pm0.0110}}$ \\
\bottomrule
\end{tabular}%
\caption{HPSv3, Aesthetic, and ImageReward results on the shared four-suite prompt panel.
\ours improves all three metrics on every prompt suite.}
\label{tab:supp-external-quality}
\end{table}

\begin{table}[H]
\centering
\small
\setlength{\tabcolsep}{6pt}
\begin{tabular}{@{}lrrrr@{}}
\toprule
Constraint family & Prompts & Flow-GRPO & \ours & Gain \\
\midrule
Grounding & 302 & $0.6293_{\pm0.0311}$ & $\mathbf{0.7683_{\pm0.0175}}$ & $+0.1390$ \\
Spatial & 665 & $0.2924_{\pm0.0094}$ & $\mathbf{0.4187_{\pm0.0164}}$ & $+0.1263$ \\
Size & 285 & $0.1816_{\pm0.0361}$ & $\mathbf{0.2722_{\pm0.0270}}$ & $+0.0906$ \\
Topology & 242 & $0.1525_{\pm0.0313}$ & $\mathbf{0.2385_{\pm0.0543}}$ & $+0.0860$ \\
Cardinality & 367 & $0.6790_{\pm0.0080}$ & $0.6644_{\pm0.0568}$ & $-0.0146$ \\
\bottomrule
\end{tabular}
\caption{\ours improves grounding, spatial, size, and topology constraints on VVRBench-Fast.
Each entry is the mean verifier score on the prompts that contain the constraint family, averaged within each training seed and reported as mean and sample standard deviation across the three seeds.
Bold marks gains whose displayed intervals do not overlap.}
\label{tab:supp-vvr-family}
\end{table}

\section{Latent Channel Analysis}
\label{sec:supp-latent-channel}

The analysis uses the seed 1 \ours denoiser and $g_\phi$ at the end of training, together with the first 100 prompts in the fixed GenEval2 training manifest.
For each prompt, we sample one initial latent and generate a reference image with the deterministic flow-matching ODE.
At every one of the 25 denoising steps, we draw a spatial Gaussian field, normalize it to zero mean and unit standard deviation, and add it to one selected latent channel with either positive or negative sign.
The perturbation uses the Flow-GRPO noise coefficient $a_t$ with $\lambda=0.7$.
All other channels remain on the deterministic trajectory.
Repeating this procedure for 16 channels gives 1,600 antithetic pairs, or 3,200 perturbed generations.

We evaluate each perturbed image by its mean absolute RGB difference from the reference.
For the same prompt, \ours is evaluated along the unperturbed reference trajectory.
Its predicted log scale is averaged over spatial positions and then over denoising steps for each channel.
The two signs are averaged to obtain one visual change value per prompt and channel.

Table~\ref{tab:latent-channel-calibration} reports the correlations at three aggregation levels.
The four channels assigned the largest scales are 14, 2, 7, and 13 and average $0.0812$ RGB change.
The four channels assigned the smallest scales are 12, 4, 0, and 10 and average $0.0570$.

\begin{table}[H]
\centering
\small
\setlength{\tabcolsep}{3pt}
\begin{tabular}{@{}lrrrr@{}}
\toprule
Aggregation & $N$ & Pearson $r$ & Spearman $\rho$ & 95\% Pearson CI \\
\midrule
Prompt and channel combinations & 1600 & $+0.306$ & $+0.332$ & $[+0.273,+0.342]$ \\
Centered within prompt & 1600 & $+0.461$ & $+0.490$ & $[+0.438,+0.485]$ \\
Channel means & 16 & $+0.598$ & $+0.491$ & $[+0.145,+0.844]$ \\
Within prompt, then averaged & 100 & $+0.485$ & $+0.490$ & $[+0.462,+0.506]$ \\
\bottomrule
\end{tabular}
\caption{\ours's predicted scale tracks intervention sensitivity across prompt and channel aggregation levels.
The first two intervals and the last resample prompts with replacement over 20,000 replicates; the channel-means interval is a Fisher interval over the 16 channels.
The within-prompt correlation is positive for all 100 prompts, with minimum Spearman $\rho=+0.156$.}
\label{tab:latent-channel-calibration}
\end{table}

\paragraph{Human perceptual evaluation.} The human study uses prompts at fixed manifest indices 0, 20, 26, 47, 67, and 98 from the GenEval2 training set and the pretrained, Flow-GRPO, and \ours denoisers.
For every model and prompt, we generate a deterministic reference and intervene on each of the 16 latent channels with both signs.
This gives 288 trials, each containing a reference and two perturbed images, and 576 perturbed images in total.
Model, channel, and perturbation sign are hidden from annotators.
Nine annotators participated, and each trial contributes exactly three independent ratings.
When four ratings were available, we dropped one at random for each trial using a fixed seed before aggregation.
Each image receives an ordinal change magnitude rating from 0 for no obvious visible difference to 4 for a very large difference.

We use the median magnitude across annotators for each perturbed image.
Within each prompt, channels are ranked by \ours's mean predicted log standard deviation along the reference trajectory.
Table \ref{tab:human-channel-quartiles} compares the top and bottom four channels.
The pooled estimate averages both perturbation signs and all three denoisers; confidence intervals resample the six prompts over 20,000 replicates.

\begin{table}[H]
\centering
\small
\setlength{\tabcolsep}{4pt}
\begin{tabular}{@{}lrrrr@{}}
\toprule
Denoiser & Bottom quartile & Top quartile & Difference & 95\% CI \\
\midrule
Pretrained & 1.521 & 2.229 & $+0.708$ & $[+0.271,+0.958]$ \\
Flow-GRPO & 1.125 & 1.313 & $+0.188$ & $[-0.188,+0.521]$ \\
\ours & 0.771 & 1.250 & $+0.479$ & $[+0.250,+0.708]$ \\
Pooled & 1.139 & 1.597 & $+0.458$ & $[+0.319,+0.583]$ \\
\bottomrule
\end{tabular}
\caption{Cross denoiser comparison of human change ratings for \ours's largest and smallest scale quartiles.
The separation is present before training and remains after \ours training.
Quartiles are defined within each prompt, and ratings use a scale from 0 to 4.}
\label{tab:human-channel-quartiles}
\end{table}

Magnitude ratings form the perceptual analysis.
Across the twelve available comparisons between pairs of annotators, ratings are within one scale point in $84.7\%$ of cases, with mean Spearman correlation $0.565$ and mean quadratic weighted $\kappa=0.440$.

\paragraph{Pixel-space change across denoisers.} The perturbations of the human study also record the RGB change of every perturbed image, which repeats the pixel-space analysis on the pretrained and Flow-GRPO denoisers.
We use the perturbations at the strength of the main analysis, average the two signs and the six prompts for each channel, and correlate the result across the 16 channels with \ours's channel mean log scale from the main analysis.
Table~\ref{tab:pixel-channel-denoisers} also reports the rank correlation computed within each prompt, using that prompt's scales, and the number of prompts for which it is positive.
We also repeat the full pixel-space analysis on the pretrained denoiser with the 100 prompts, seeds, and perturbations of the main analysis, giving 3,200 perturbed generations.
On both prompt sets, the correlation on the pretrained denoiser is close to that on the \ours denoiser, so the relation between \ours's scales and pixel-space sensitivity is present before RL training.

\begin{table}[H]
\centering
\small
\setlength{\tabcolsep}{4pt}
\begin{tabular}{@{}lrrrrr@{}}
\toprule
Perturbed denoiser & Pearson $r$ & Spearman $\rho$ & 95\% Pearson CI & Within-prompt $\rho$ & Positive prompts \\
\midrule
\multicolumn{6}{@{}l}{\emph{100 prompts of the main analysis}} \\
Pretrained & $+0.516$ & $+0.488$ & $[+0.027,+0.806]$ & $+0.421$ & 99/100 \\
\ours & $+0.598$ & $+0.491$ & $[+0.145,+0.844]$ & $+0.490$ & 100/100 \\
\addlinespace
\multicolumn{6}{@{}l}{\emph{Six prompts of the human study}} \\
Pretrained & $+0.562$ & $+0.535$ & $[+0.092,+0.827]$ & $+0.451$ & 6/6 \\
Flow-GRPO & $+0.509$ & $+0.494$ & $[+0.018,+0.802]$ & $+0.425$ & 6/6 \\
\ours & $+0.576$ & $+0.618$ & $[+0.113,+0.834]$ & $+0.523$ & 6/6 \\
\bottomrule
\end{tabular}
\caption{Correlation between \ours's predicted scale and pixel-space change when each denoiser is perturbed.
The first three columns use channel means; intervals are Fisher intervals over the 16 channels.
Within-prompt $\rho$ averages the per-prompt rank correlation.}
\label{tab:pixel-channel-denoisers}
\end{table}